\documentclass[10pt,twocolumn,letterpaper]{article}
\pdfoutput=1
\usepackage{iccv}
\usepackage{times}
\usepackage{epsfig}
\usepackage{graphicx}
\usepackage{amsmath}
\usepackage{amssymb}

\newif\ifdraft
\draftfalse

\usepackage[normalem]{ulem}

\ifdraft
 \newcommand{\PF}[1]{{\color{red}{\bf PF: #1}}}
 
 \newcommand{\WL}[1]{{\color{blue}{\bf WL: #1}}}
 
 \newcommand{\CL}[1]{{\color{green}{\bf CL: #1}}}
 
 \newcommand{\SAM}[1]{{\color{purple}{\bf SAM: #1}}}
 
 \newcommand{\ALEX}[1]{{\color{orange}{\bf ALEX: #1}}}

\else
 \newcommand{\PF}[1]{}
 
 \newcommand{\WL}[1]{}
 
 \newcommand{\CL}[1]{}
 
 \newcommand{\SAM}[1]{}
 
 \newcommand{\ALEX}[1]{}

\fi

\newcommand{\comment}[1]{}
\newcommand{\parag}[1]{\paragraph{#1}}

\newcommand{\cF}{\mathcal{F}}

\newcommand{\f}{\mathbf{f}}

\newcommand{\x}{\mathbf{x}}
\newcommand{\bx}{\hat{\mathbf{x}}}
\newcommand{\y}{\mathbf{y}}
\newcommand{\by}{\hat{\mathbf{y}}}
\newcommand{\z}{\mathbf{z}}
\newcommand{\bz}{\hat{\mathbf{z}}}

\newcommand\norm[1]{\left\lVert#1\right\rVert}

\usepackage{multirow}

\DeclareMathOperator*{\argmax}{argmax}

\usepackage{color}

\usepackage[table]{xcolor}
\usepackage{soul}
\usepackage{booktabs}
\usepackage[normalem]{ulem}

\usepackage{caption}
\usepackage{subcaption}

\usepackage[pagebackref=true,breaklinks=true,letterpaper=true,colorlinks,bookmarks=false]{hyperref}

\iccvfinalcopy 

\begin{document}

\title{Domain Adaptation for Semantic Segmentation \\ via Patch-Wise Contrastive Learning}

\author{
	Weizhe Liu\textsuperscript{1}
	\quad
	David Ferstl \textsuperscript{2}
	\quad
    Samuel Schulter \textsuperscript{3}
    \quad
    Lukas Zebedin  \textsuperscript{3}
    \quad
	Pascal Fua\textsuperscript{1}
    \quad 
    Christian Leistner\textsuperscript{3} \\
	\textsuperscript{1} CVLab, EPFL 
    \quad \textsuperscript{2} Magic Leap 
    \quad \textsuperscript{3} Amazon \\
}

\maketitle
\begin{abstract}

We introduce a novel approach to unsupervised and semi-supervised domain adaptation for semantic segmentation. Unlike many earlier methods that rely on adversarial learning for feature alignment, we leverage contrastive learning to bridge the domain gap by aligning the features of structurally similar label patches across domains. 
As a result, the networks are easier to train and deliver better performance. Our approach consistently outperforms state-of-the-art unsupervised and semi-supervised methods on two challenging domain adaptive segmentation tasks, particularly with a small number of target domain annotations.  It can also be naturally extended to weakly-supervised domain adaptation, where only a minor drop in accuracy can save up to $75\%$ of annotation cost.

\end{abstract}
\section{Introduction}

Given large amounts of annotated training data, current fully-supervised semantic segmentation algorithms deliver outstanding results. Because annotating many images at the pixel-level is expensive, a common practice is to generate synthetic data and to rely on unsupervised domain adaptation to bridge the gap from the synthetic source to the real-world target domain. 

In this paper, we introduce a novel approach to aligning cross-domain features for both unsupervised and semi-supervised domain adaptation. In the first case, no target domain labels are given, whereas in the second one only a small amount of annotated data is available in the target domain. In practice, this second scenario is important because even a handful of target domain labels can boost the performance significantly.


\begin{figure}[t]
\centering
\includegraphics[width=1.0\linewidth]{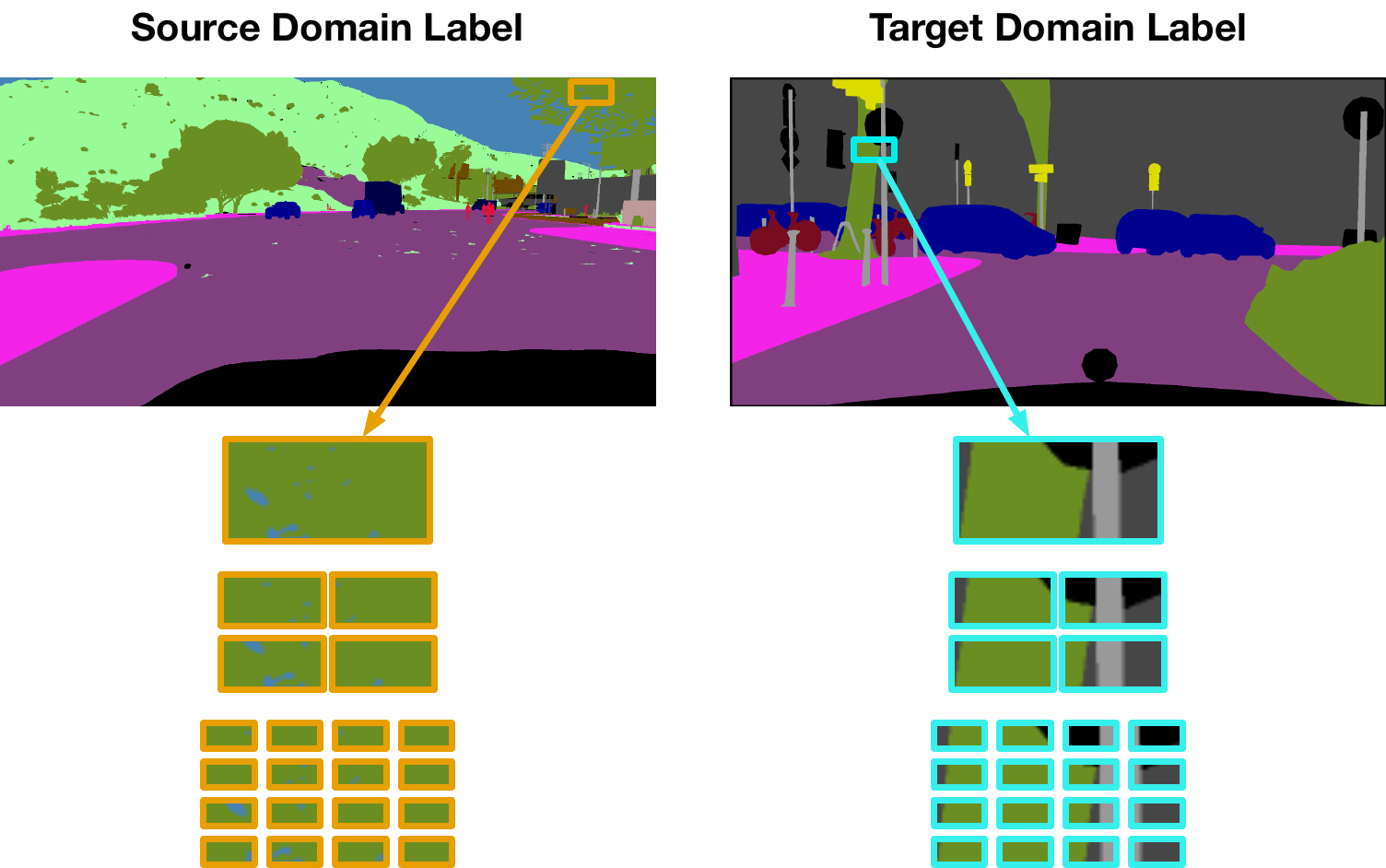}
\caption{\small {\bf Patch-wise structural disparity.} Patches from both domains are divided into  sub-patches at several resolution levels. We then compare their label space disparity at all levels and use contrastive learning to make similar patches have similar feature distributions.  }
\label{fig:intro}
\end{figure}

Our key insight is that patches from both domains that are {\it structurally} similar in label space should also have similar distributions in feature space. To enforce this, which goes beyond what can be done using ordinary pixel-wise similarity, we introduce a patch-wise metric. It measures label disparity at several levels of resolution along with a contrastive loss~\cite{Chen20} that, when minimized, aligns the feature distribution closer for patches with similar structures in label space and pushes them apart otherwise, see Fig.~\ref{fig:intro}. 

To perform unsupervised or semi-supervised training, we incorporate unlabeled pixels into training by using {\it pseudo labels} that we iteratively generate using the output of the partially trained network. At each training iteration, we use our patch-wise metric to decrease the feature space disparity of patches that are structurally similar and to increase it for those that are not. This renders a more straightforward approach than adversarial learning, previously often used for cross-domain feature alignment. We do not require an extra discriminator network and therefore eliminate the sometimes substantial difficulty of having to train it.

Our experiments show that our approach delivers a performance increase over state-of-the-art methods in the unsupervised regime~\cite{Vu19,Yang20a} and an even larger boost in the semi-supervised one~\cite{Wang20,Kalluri19}. In a practical setting, we believe the latter to be particularly significant because, while it is very tedious to get huge amounts of annotated frames, it is almost always possible to supply {\it a few} annotated frames. 
Our contribution is therefore a new contrastive learning approach to aligning features across different domains for semantic segmentation. It relies on structural label disparity instead of adversarial learning and outperforms state-of-the-art methods for unsupervised domain adaptation and even more for semi-supervised domain adaptation. We also show how our approach can be extended to weakly-supervised domain adaptation, where only a minor drop in accuracy can save up to $75\%$ of annotation cost.

\section{Related Work}

Most recent works on domain adaptation focus on unsupervised methods, with only very few incorporating the limited amount of supervision we advocate. We briefly review these two classes of approaches and then discuss contrastive learning, which is a central component of our approach.

\parag{Unsupervised Domain Adaptation (UDA) for Semantic Segmentation} aims to align the source and target domain feature distributions given annotated data {\it only} in the source domain. 
A popular approach is to leverage adversarial learning to generate domain-invariant features.
This trend started with~\cite{Hoffman16} and was extended to different levels of representation, including feature space~\cite{chen18,Hong18,Paul20,Saito18a} and label space~\cite{Chen17,Huang20,Pan20,Tsai18,Vu19,Vu19b}. Notable extensions are~\cite{Chen19,Chen17,Paul20} which enforce class-wise alignment to narrow the distributions to be matched. However, all these adversarial learning approaches rely on extra discriminator networks which are complicated and hard to train jointly with the generator network.

Other widely used UDA approaches to semantic segmentation include generating realistic-looking synthetic images~\cite{Hoffman18,Sankaranarayanan18,Wu18,Yang20b,Yang20a,Zhu18}, using pseudo labels for self-training~\cite{Li20,Lian19,Subhani20,Yang20a,Zou18}, and leveraging weak labels~\cite{Lv20,Paul20}. Among these methods, FDA~\cite{Yang20a} is a simple approach which achieves state-of-the-art performance by generating realistic-looking synthetic images directly in Fourier space and self-training.

Our approach builds on FDA and incorporates a novel domain-wise contrastive loss, without the reliance on complex adversarial learning.

\parag{Semi-Supervised Domain Adaptation for Semantic Segmentation (SSDA)} 
assumes that a handful of target domain labels are available. In the context of semantic segmentation it has not received as much attention as UDA. For example, \cite{Saito19} achieves adaptation by alternately maximizing the conditional entropy of unlabeled target data with respect to the classifier and minimizing it with respect to the feature encoder. In~\cite{Rozantsev19}, a two-stream architecture is proposed, where one stream operates in the source and the other in the target domain. In contrast to others, the weights in corresponding layers are related but not shared and optimized to deliver good performance in both domains. In~\cite{KimT20}, target domain samples are perturbed to reduce intra-domain discrepancy using adversarial learning.

However, none of these works have been demonstrated for semantic segmentation, nor do they leverage contrastive learning for SSDA. The only ones that do are~\cite{Wang20} and~\cite{Kalluri19}. In~\cite{Wang20}, class-wise adversarial learning is used to promote the similarity of  pixel-level feature representations for the same classes. In~\cite{Kalluri19}, a pixel-level entropy regularization scheme is introduced to favor feature alignment among multiple domains. Therefore, domain alignment is only enforced on the pixel-level, whereas ours is done at a more semantic level.

\parag{Contrastive Learning (CL)} aims to learn visual representations by leveraging both similar and dissimilar samples. Early work~\cite{Hadsell06} showcased improved visual representation by contrasting positive pairs against negative ones. Deep CL has been used extensively~\cite{Dosovitskiy14, Wu18b, Zhuang19, Tian20, He20, Misra20, Chen20, Li20b, Doersch17, Ye19, Ji19, khosla2020} for applications such as image classification, image-to-image translation~\cite{Park20} or phrase grounding~\cite{Gupta20}. 

In the context of semantic segmentation, CL has also been used for intra-domain  model pre-training ~\cite{Chaitanya20}. The approach is specifically designed for use in conjunction with a U-Net~\cite{Ronneberger15} backbone and does not generalize well to other architectures. In our experiments, we show that using CL directly as loss term achieves superior results compared to CL only used for model pre-training.

\section{Approach}
\label{fig:approach}

We first formalize the problem of domain adaptation in Sec.~\ref{sec:problem_formulation}.  We then give an overview of our proposed model in Sec.~\ref{sec:model_overview} and explain how contrastive learning is used for domain adaptation in Sec.~\ref{sec:constrastive} and \ref{sec:training}.

\subsection{Problem Formulation}
\label{sec:problem_formulation}

Let $D^{s} = \{(\x^{s}_{i},\y^{s}_{i})\}_{i=1}^{N_{s}}$ be a source-domain dataset, where $\x^{s}$ denotes a color image and $\y^{s}$ the corresponding semantic map. The target-domain dataset is split into two sets.  The first is a labeled set $D^{l} = \{(\x^{l}_{i},\y^{l})\}^{N_{l}}_{i=1}$ with ground truth semantic maps.  The second is the unlabeled set $D^{u} = \{\x^{u}_{i}\}^{N_{u}}_{i=1}$ without ground truth semantic labels. We have $N_{l}=0$ in the Unsupervised Domain Adaptation (UDA) setting and $N_{l}>0$ for Semi-Supervised Domain Adaptation (SSDA). In most real-world scenarios, we have $N_{u} \gg N_{l}$ and $N_{s} \gg N_{l}$.  Given the three data sets $D^{s}$, $D^{l}$ and $D^{u}$, the task is to learn a single model that performs well on previously unseen target domain data.


\begin{figure*}[t]
\centering
\includegraphics[width=1.0\linewidth]{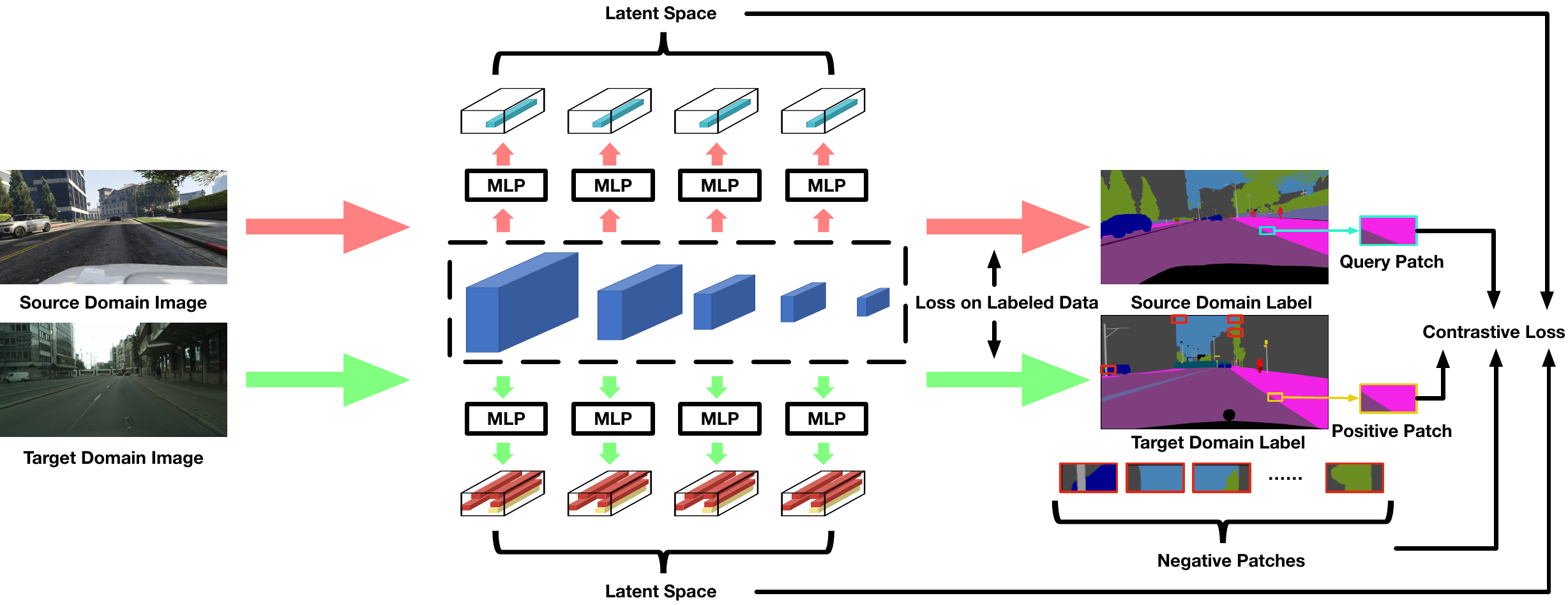}
\vspace{-6mm}
\caption{\small {\bf Model Architecture.}  Images from both domains are fed into the same deep network. Features after each convolutional block of ResNet101 are projected into the latent space by a MLP where the contrastive loss is employed. The contrastive loss minimizes the distance of the latent space between query patch and positive patch, while maximizing the distance for the query patch and negative patches.}
\label{fig:model}
\end{figure*}

\subsection{Overview of Our Approach}
\label{sec:model_overview}

Our model consists of an encoder ${\cal E}$ that maps an input image $\x$ to a list of feature vectors $\f={\cal E}(\x)$ and two decoders ${\cal D}_{la}$ and ${\cal D}_{se}$  that map these intermediate features to a list of latent vectors $\z={\cal D}_{la}(\f)$ and a semantic map $\y={\cal D}_{se}(\f)$. Let ${\cal F}_{se} = {\cal D}_{se} \circ {\cal E}$ and ${\cal F}_{la} = {\cal D}_{la} \circ {\cal E}$ be the networks that take as input an image and return a semantic map and a latent vector respectively.
Fig.~\ref{fig:model} gives an overview of our network architecture and the loss functions, which we describe below.

\parag{Patch-wise representation:}
In practice, the functions ${\cal E}$ and ${\cal D}_{se}$ are convolutional neural networks, but for the sake of describing our contrastive loss in Sec.~\ref{sec:constrastive}, a patch-wise representation is useful.
Therefore, each image $\x$ is decoded into $N_p$ latent vectors, corresponding to $N_p$ rectangular patches $\bx_i$ for $1 \leq i \leq N_p$ and the encoder preserves the association between features and patches so that we can write
\begin{align}
\f ={\cal E}(\x)&= {\cal E}\left(\left[\bx_1,\ldots,\bx_{N_p}\right]\right) = \left[\hat{\f}_1,\ldots,\hat{\f}_{N_p}\right] \label{eq:features} \\
\z ={\cal F}_{la}(\x)& = {\cal D}_{la}(\f)= {\cal D}_{la}\left(\left[\hat{\f}_{1},\ldots,\hat{\f}_{N_p}\right]\right) \nonumber \\
& = \left[\bz_1,\ldots,\bz_{N_p}\right] \nonumber
\end{align}
where $\hat{\f}_i$ and $\bz_i$ are local features and latent vectors associated to patch $\bx_i$.
In the remainder of the paper, we will denote by $\bx^s$, $\bx^l$ and $\bx^u$ patches of the source, labeled and unlabeled target images.

\parag{Loss functions for classification:}
Training all parameters of our model involves minimizing several loss functions, including the novel contrastive loss that we describe in Sec.~\ref{sec:constrastive}.  For the per-pixel classification task, we also use supervised cross-entropy losses for the source domain images and, if available, labeled target domain images, respectively, which are defined as
\begin{align}
    L_{sup}^{s}  & = -\sum_{i}\left \langle \y_{i}^{s},log({\cal F}_{se}(\x_{i}^{s})) \right \rangle \;, \label{eq:loss_sup_source} \\
    L_{sup}^{l} &= -\sum_{i} \left \langle \y_{i}^{l},log({\cal F}_{se}(\x_{i}^{l}))\right \rangle \;. \label{eq:loss_sup_target}
\end{align}
Additionally, we employ a regularization loss 
\begin{equation}
    L_{ent} = \sum_{i}\rho\left(- \left \langle {\cal F}_{se}(\x_{i}^{u}),log({\cal F}_{se}(\x_{i}^{u}))\right \rangle\right)\;, \label{eq:loss_ent}
\end{equation}
where $\rho(x) = (x^{2}+0.001^{2})^{\eta}$ is the Charbonnier penalty function~\cite{Bruhn05}.  As in~\cite{Yang20a}, $L_{ent}$ penalizes uncertainty in the predictions for the unlabeled samples and encourages one label to dominate over the others.  

We summarize these three losses into our base loss
\begin{equation}
    L_{base} = L_{sup}^{s}+ L_{sup}^{l}+ \lambda_{ent} L_{ent}\;, \label{eq:loss_base}
\end{equation}
with $\lambda_{ent}$ being a small scaling factor, usually $<1$.

\parag{Pseudo Labels:} After some initial training steps, we can use ${\cal F}_{se}$ to assign \emph{pseudo labels} to unlabeled images and to compute an additional cross-entropy loss term
\begin{equation}
    L_{self} = -\sum_{i}\left \langle \y_{i}^{u},log(\cF_
    {se}(\x_{i}^{u}))\right \rangle \;, \label{eq:loss_self}
\end{equation}
where $\y_{i}^{u}=\argmax({\cal F}_{se}(\x_{i}^{u}))$ are the pseudo labels.
In Sec.~\ref{sec:ablation}, we demonstrate that pseudo labeling becomes more effective in the SSDA setting, as compared to an unsupervised setting, because pseudo labels are more reliable.

\subsection{Contrastive Learning for Domain Adaptation}
\label{sec:constrastive} 
The main contribution of this paper is to leverage contrastive learning for domain adaptation.
Instead of relying on an adversarial training scheme, as most prior works do~\cite{Wang20,Vu19}, we use a contrastive loss on pairs of patches from different domains.
The goal is to bring the representation of positive pairs closer together, while pushing negative pairs apart.
A benefit of our approach is that optimizing this loss is relatively simpler, compared to adversarial learning, which involves a min-max optimization scheme.
However, the main challenge is to define positive and negative pairs of patches across domains for the contrastive loss.

\subsubsection{Matching of Patches for Contrastive Learning}
\label{sec:finding_pairs}

The key idea of our approach is that if two patches (one from the source domain and the other from the target domain) are semantically similar, then  their embedding in latent space $\bz_i = {\cal F}_{la}(\bx_{i})$ should also be similar.  Conversely, if two patches are semantically dissimilar, their embeddings should also have a large distance.

\begin{figure}[t]
\centering
\includegraphics[width=1.0\linewidth]{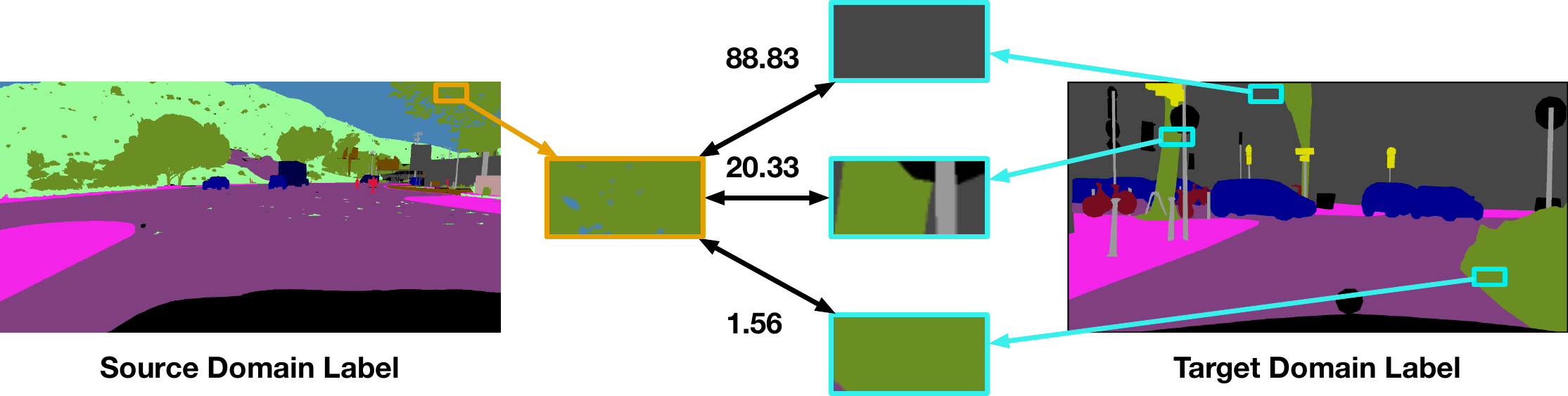}
\vspace{-6mm}
\caption{\small {\bf Example of Semantic Disparity.}  We visualize the semantic disparity between one label patch from the source domain and three label patches from the target domain. As can be seen, our metric clearly demonstrates both semantic and spatial disparity.}
\label{fig:matching}
\end{figure}

To find such pairs, let us assume a semantic {\it disparity} function $D$, which we define formally in the following section. Patches with high semantic similarity have low disparity values $D$, and vice-versa.  We sample a patch in an image of one domain and compute the disparity to all patches in an image from the other domain.  Pairs of patches with low disparity score are considered positive ($D < \alpha$) and pairs with high disparity negative ($D > \beta$). Pairs with disparity values in-between $\alpha$ and $\beta$ are simply ignored.
Fig.~\ref{fig:matching} gives an example of positive and negative pairs.

To define a contrastive loss with the discovered pairs of patches, let us define the \emph{query patch} $\bx^q$ sampled from one domain, the \emph{positive patch} $\bx^+$ and $k$ \emph{negative patches} $\{\bx^-_{i}\}_{i=1}^{k}$ sampled from the other domain.  Let $\bz^q$, $\bz^+$ and $\bz^-_{i}$ be the corresponding latent vectors.  We can then define the \emph{contrastive loss} as
\begin{small}
\begin{equation}
    L_{cont} =  -\log\left[\frac{sim(\bz^q,\bz^{+})}{sim(\bz^q,\bz^{+})+\sum_{i=1}^{k} sim(\bz^q,\bz^{-}_{i})}\right]  \;,
    \label{eq:cont}
\end{equation}
\end{small}%
with $sim(u,v) = \exp\left(\frac{u^{T}v}{\norm{u}\norm{v}\tau}\right)$ being the similarity between any two vectors, defined as the exponential of the cosine similarity normalized by temperature parameter $\tau$.


\subsubsection{Patch-Wise Semantic Disparity}
\label{sec:patch_similarity}

We need a measure of patch-wise semantic disparity in label space to define positive and negative pairs as described in Sec.~\ref{sec:finding_pairs}.
Given a pair of patches, one from source and the other from the target domain, we define a metric accounting for both  \emph{semantic} and \emph{structural} disparity in label space.


\begin{figure}[t]
\centering
\includegraphics[width=1.0\linewidth]{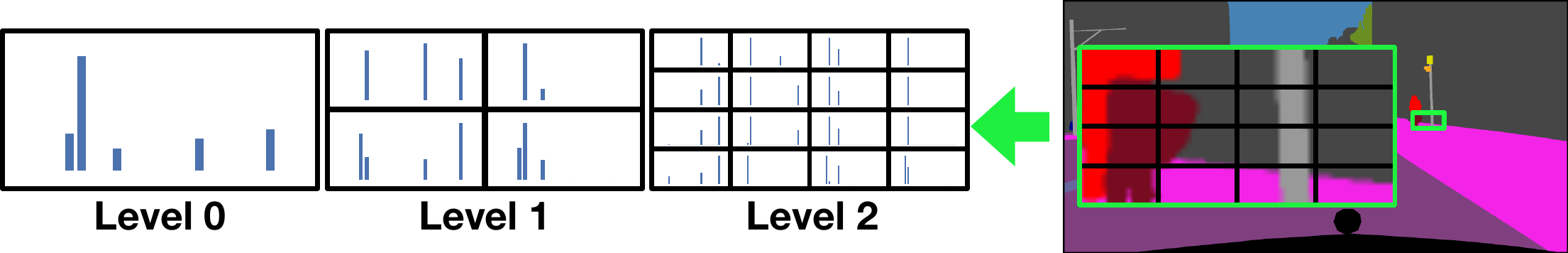}
\vspace{-6mm}
\caption{\small {\bf Label Space Spatial Pyramid Matching.} A patch is cropped from the full label mask and divided into 3 different levels, from coarse to fine. The semantic information is calculated as the normalized histograms over categories for each sub-patch in all three levels.
}
\label{fig:pyramid}
\end{figure}

Let us consider patch $\bx$ and semantic map $\by$ from either domain. Let $f_{\bx}^c$ be the proportion of pixels that have label $c$, for $1 \leq c \leq N_c$. We take
\begin{equation}
V_{\bx} = \left[f_{\bx}^{1},f_{\bx}^{2},...,f_{\bx}^{N_{c}}\right] \in \mathbb{R}^{N_{c}}
\label{eq:vector}
\end{equation}
to be the semantic vector of $\bx$, containing the overall semantic information without any spatial layout information.

To also encode rough spatial information and allow for robust matching, we adopt a simplified version of spatial pyramid matching~\cite{Lazebnik06} in label space.
We compute the semantic vector via Eq.~\eqref{eq:vector} on three spatial levels, as shown in Fig.~\ref{fig:pyramid}.
We define the \emph{patch-wise semantic disparity} between $\bx_s$ and $\bx_t$, from source and target domains, respectively, as
\begin{align}
D(\bx^{s},\bx^{t}) = \sum_{m=0}^{2} N_{2-m} D_{m}\left(\bx^{s},\bx^{t}\right)\;,
\label{eq:similarity}
\end{align}
with $D_{m}(\bx^{s},\bx^{t}) = \sum_{i=1}^{N_{m}}\norm{V_{\bx^{s}_{i,m}}-V_{\bx^{t}_{i,m}}}^{2}$.  That is, we measure $\ell_2$ distance between semantic vectors at 3 pyramid levels $m$, each with $N_{m}$ patches.
$\bx^{s}_{i,m}$ and $\bx^{t}_{i,m}$ denote the $i$-th sub-patch of $\bx^{s}$ and $\bx^{t}$ at level $m$.  The first spatial level covers the whole patch, hence $N_{0}$=1.

At the second and third levels, we split the patch into 4 and 16 sub-patches, respectively, and set $N_{1}=4$ and $N_{2}=16$.
The coefficient $N_{2-m}$ in Eq.~\eqref{eq:similarity} ensures equal contribution from all levels.

\subsection{Training Strategy}
\label{sec:training}
We can now put the individual losses Eq.~\eqref{eq:loss_base}, \eqref{eq:loss_self} and \eqref{eq:cont} together to define our overall training objective
\begin{equation}
    L_{all} = L_{base} + \lambda_{self} L_{self} + \lambda_{cont} L_{cont} \;,
    \label{eq:loss_all}
\end{equation}
which we minimize with respect to the network parameters.  We introduce weighting factors $\lambda_{self}$ and $\lambda_{cont}$ to balance the impact of individual loss terms.

We first train a network only with $L_{base}$, which we use to estimate pseudo labels.  Then, our network is re-initialized and re-trained with the full loss $L_{all}$. The contrastive loss operates on both labeled target and pseudo-labeled target data, where we use a lower weight $\lambda_{cont}$ for the one operating on pseudo-labeled data. Note again that our training approach does not require adversarial learning objectives for domain adaptation.

\subsection{Implementation Details}

We use DeepLabV2~\cite{Chen17b} as our semantic segmentation network ${\cal F}_{se}$, with the encoder ${\cal E}$ being a ResNet101~\cite{He16} backbone.  This choice enables a fair comparison with prior works~\cite{Wang20,Vu19,Yang20a}.
The decoder network ${\cal D}_{la}$, which extracts the latent variables for our contrastive loss, is more sophisticated.  It uses an average pooling layer followed by a two-layer perceptron to project the feature patches $\hat{\f}_p$ of the encoder ${\cal E}$ (see Eq.~\eqref{eq:features}) into the latent space $\bz$ in which the contrastive loss Eq.~\eqref{eq:cont} is computed. We add decoder ${\cal D}_{la}$ to multiple intermediate layers as suggested by previous work~\cite{Park20}. Note that ${\cal D}_{la}$ is only required at training time.

To improve our overall adaptation quality, we employ the recently proposed Fourier Domain Adaptation (FDA) method~\cite{Yang20a}. It translates source domain images to the target domain by swapping the low-frequency component of the spectrum of the source image with that of a randomly selected target one.  The strength of this approach is that the translation is very simple as it happens directly in the image space without any deep neural network.
Thus, the source domain images we use in Eq.~\eqref{eq:loss_sup_source} are translated using FDA. 

We use the following hyper-parameters:
As in~\cite{Yang20a}, we set $\lambda_{ent}$ to $0.005$ and $\lambda_{self}$ to $1.0$.  We then tested different values of $\lambda_{cont}$ in Eq.~\eqref{eq:loss_all}.  We set it to $1e^{-3}$ for target labels sampled from ground truth annotation and to $1e^{-4}$ for pseudo labels as mentioned in Sec.~\ref{sec:training}.

The temperature parameter $\tau$ in Eq.~\eqref{eq:cont} is set to $0.07$ during training and the patch size for the contrastive loss is set to $64 \times 32$ pixels in image space. We use $\alpha=3$ and $\beta=70$ as thresholds to define positive and negative patch pairs. Our model is trained using SGD with initial learning rate $2.5e^{-4}$ and adjusted according to the `poly' learning rate scheduler with a power of $0.9$ and weight decay $0.0005$, following~\cite{Yang20a}.
\newcommand{\ours}[0]{{\bf OURS}}
\newcommand{\oursU}[0]{{\bf OURS-UDA}}
\newcommand{\oursS}[0]{{\bf OURS-SUP}}
\newcommand{\oursB}[0]{{\bf OURS-BASE}}
\newcommand{\oursP}[0]{{\bf OURS-PRE}}
\newcommand{\udaB}[0]{{\bf UDA-BASE}}

\section{Experiments}

\subsection{Baselines}

We evaluate our proposed method on both unsupervised (UDA) and semi-supervised (SSDA) domain adaptation.  ASS~\cite{Wang20} is the only semantic segmentation work we know of that operates in the SSDA setting, \ie, assumes full annotation in the source domain and partial annotations in target domain.  For a fair comparison against other state-of-the-art methods, we extend the following baselines to handle both UDA and SSDA setups:
\begin{itemize}

    \item FDA~\cite{Yang20a} translates source domain images to the target style by swapping low-frequency values in Fourier space and leveraging self-training to refine the estimation.
    
    \item MinEnt~\cite{Vu19} minimizes an entropy loss to penalize low-confidence predictions in the target domain. 
    
    \item AdvEnt~\cite{Vu19} minimizes the same entropy loss as MinEnt and also performs structure adaptation from source to target domain in an adversarial setting.
    
    \item Universal~\cite{Kalluri19} introduces a pixel-level entropy regularization scheme to perform feature alignment among multiple domains. 
    
\end{itemize}
The first three were designed for UDA while the fourth performs SSDA but assumes partial annotations in both domains.

To also test FDA, MinEnt, and AdvEnt in a semi-supervised context we modified them to also leverage annotated target domain data. To this end, we train them by minimizing their original objective function along with the loss from Eq.~\eqref{eq:loss_sup_target} for supervised target domain data. For Universal, we used all the source domain annotations, replaced the original network by the same DeepLabV2~\cite{Chen17b} network as for all the other baselines and kept the rest of the method as in the original work, which boosted its performance and makes the comparison fair.


\begin{figure*}[t]
\centering
\begin{tabular}{cccc}
\includegraphics[width=.22\linewidth]{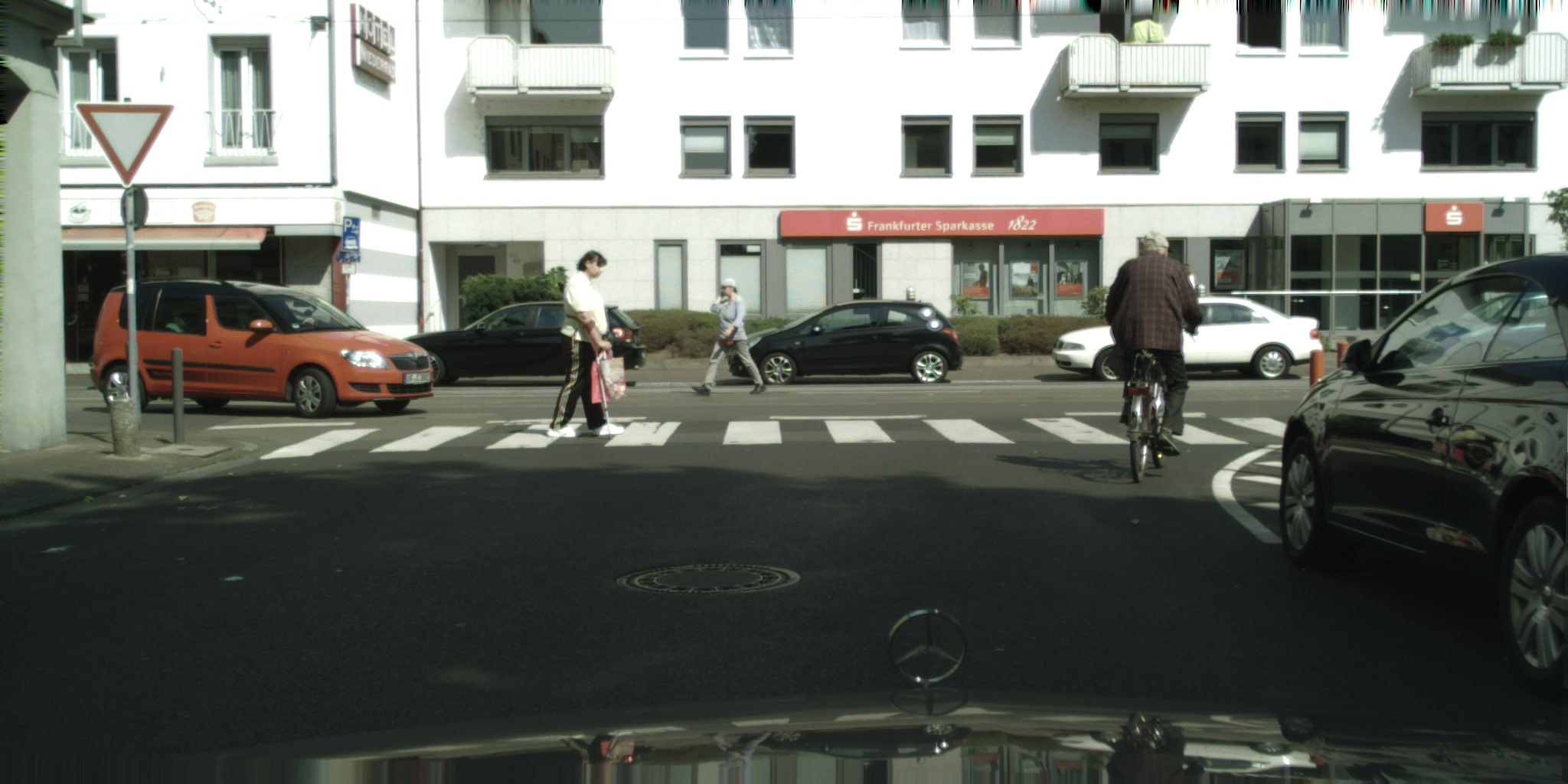}&
\includegraphics[width=.22\linewidth]{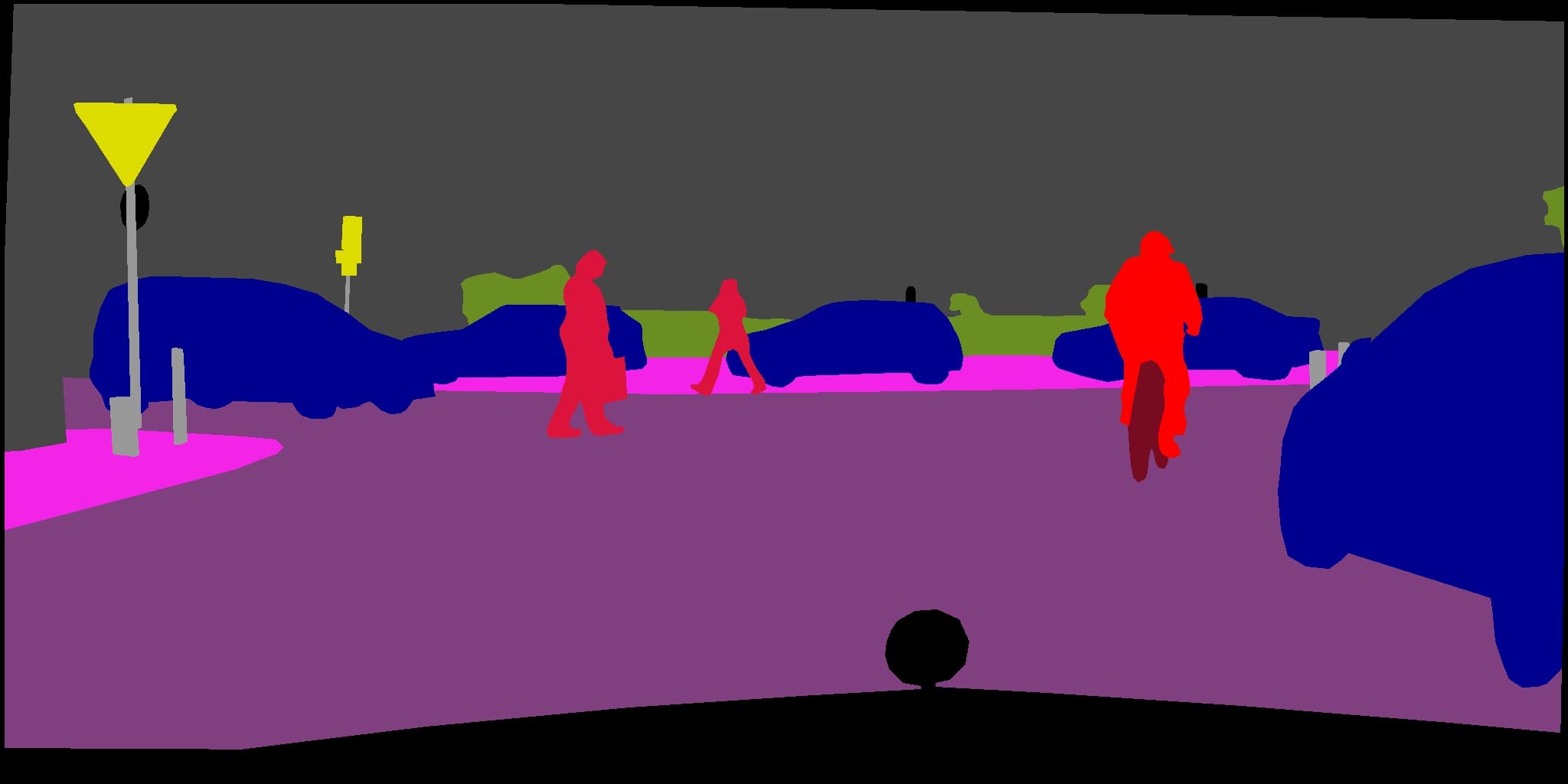}&
\includegraphics[width=.22\linewidth]{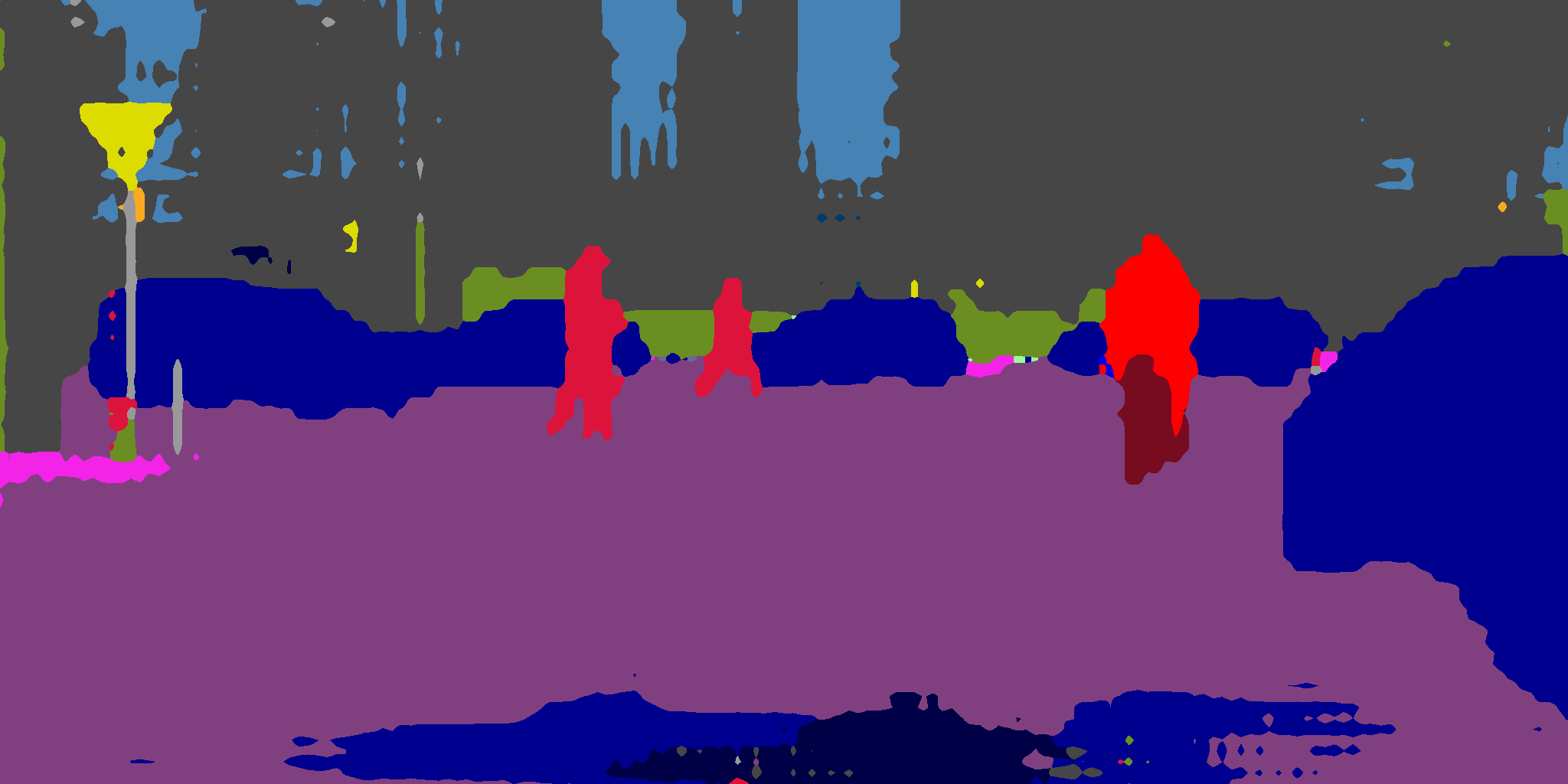}&
\includegraphics[width=.22\linewidth]{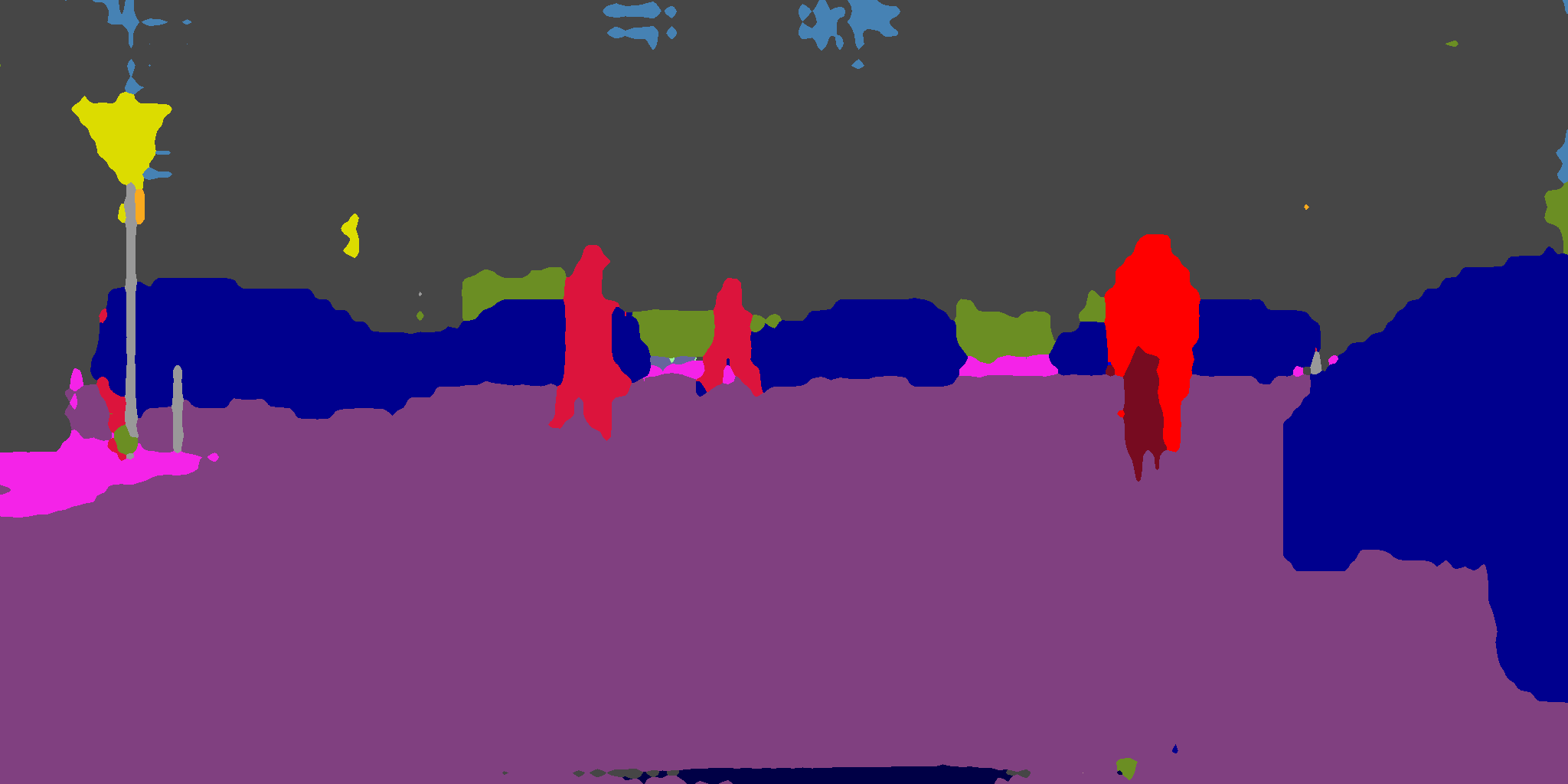}\\
\includegraphics[width=.22\linewidth]{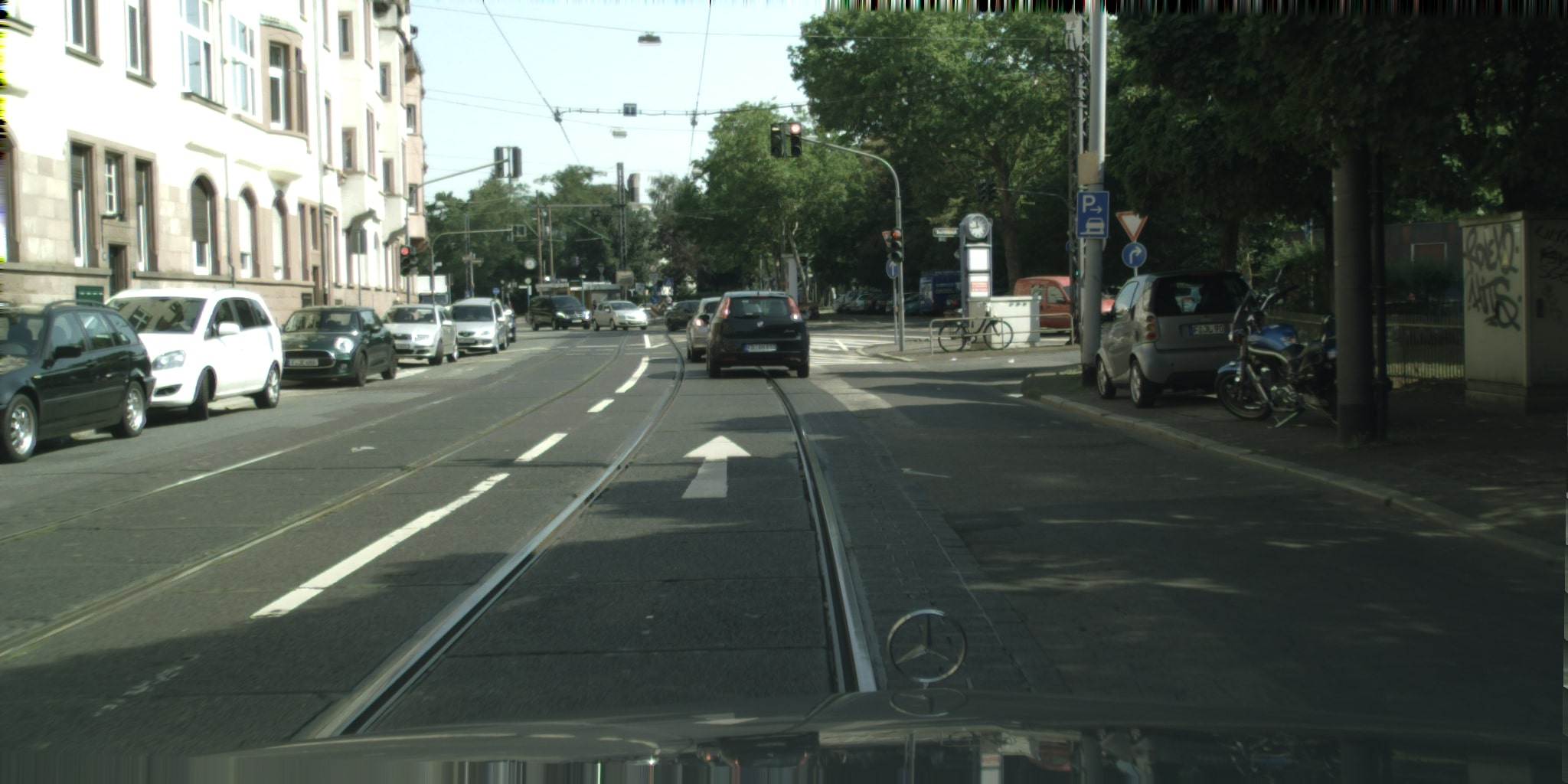}&
\includegraphics[width=.22\linewidth]{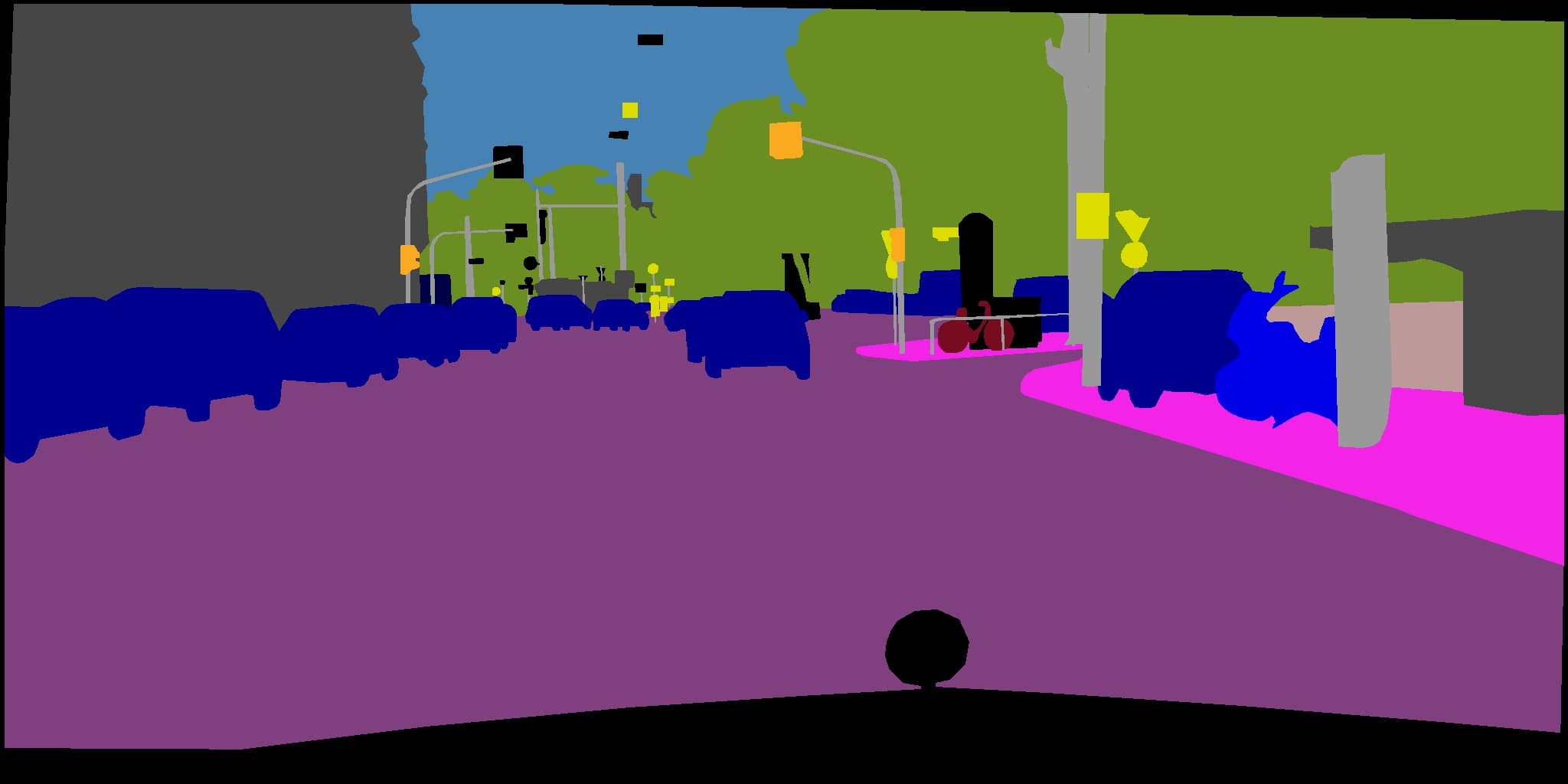}&
\includegraphics[width=.22\linewidth]{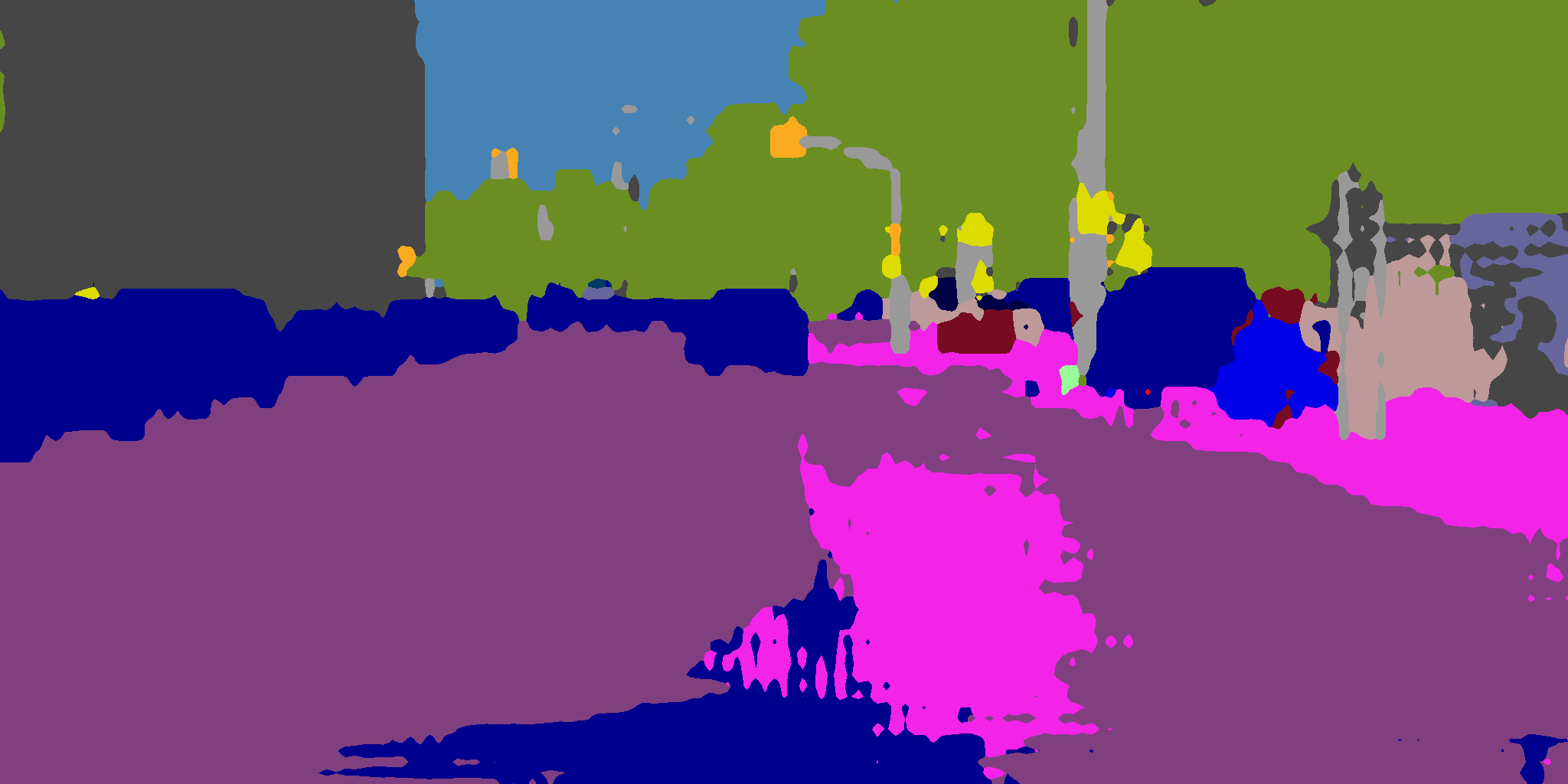}&
\includegraphics[width=.22\linewidth]{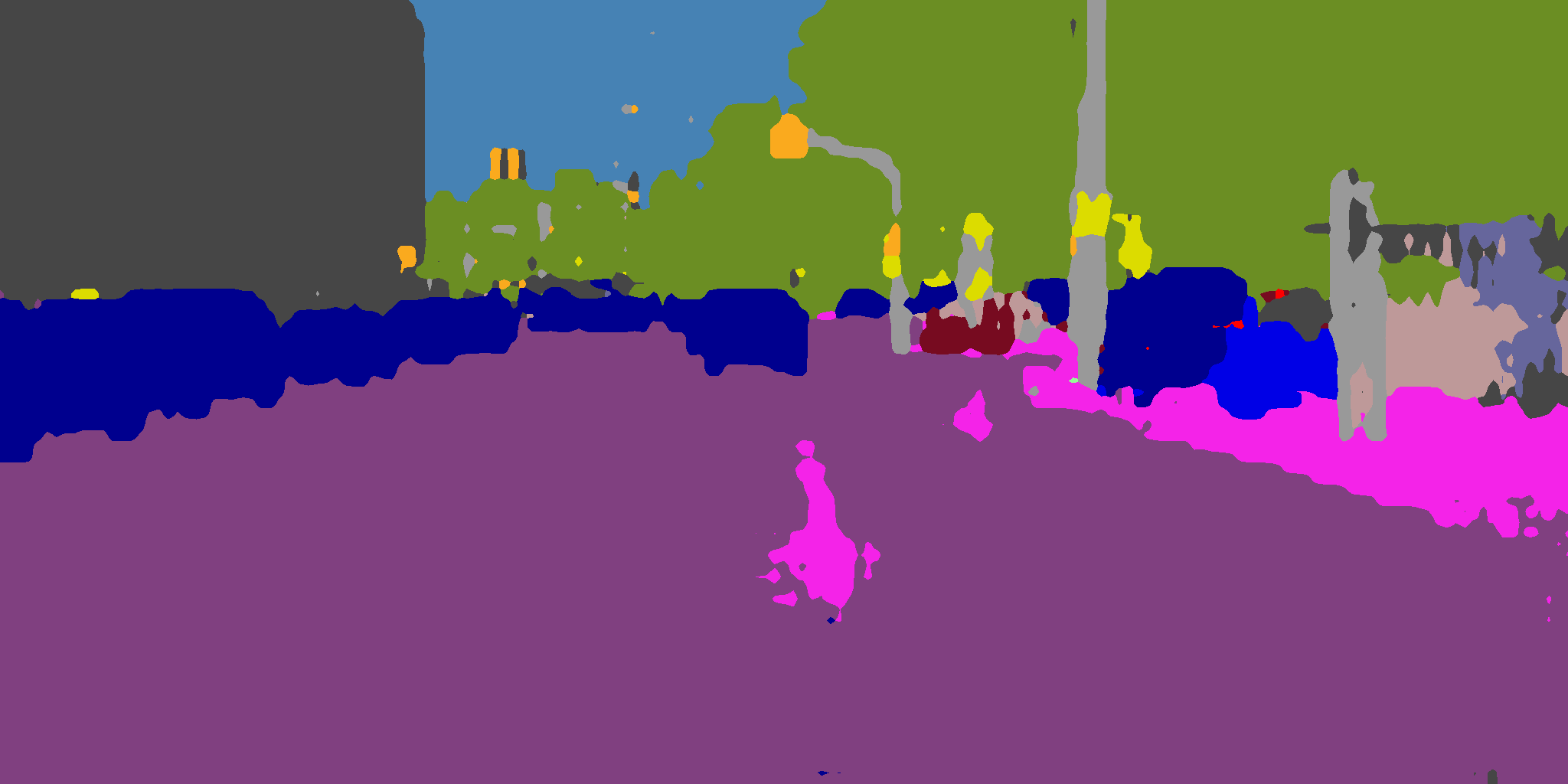}\\
\includegraphics[width=.22\linewidth]{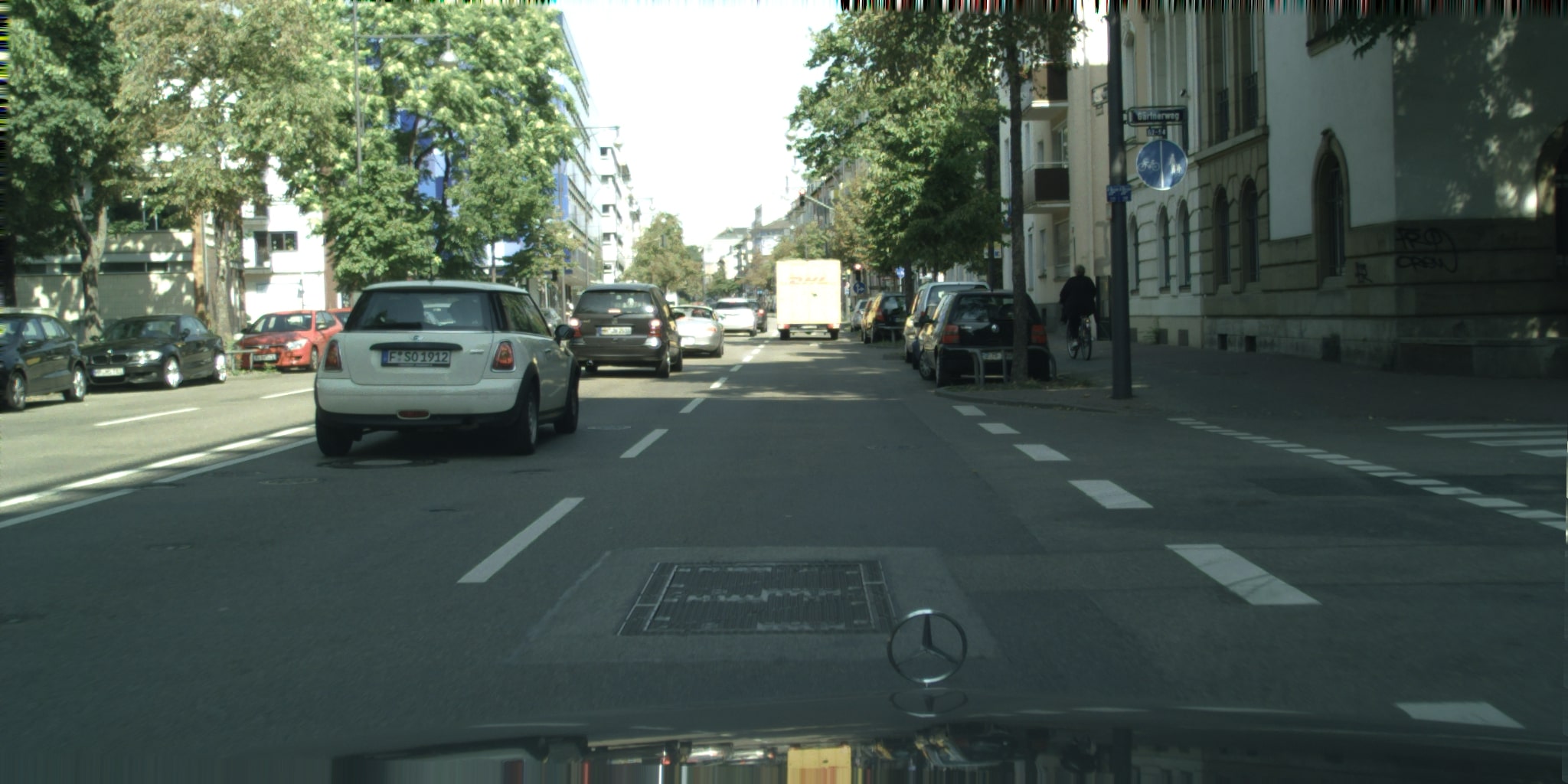}&
\includegraphics[width=.22\linewidth]{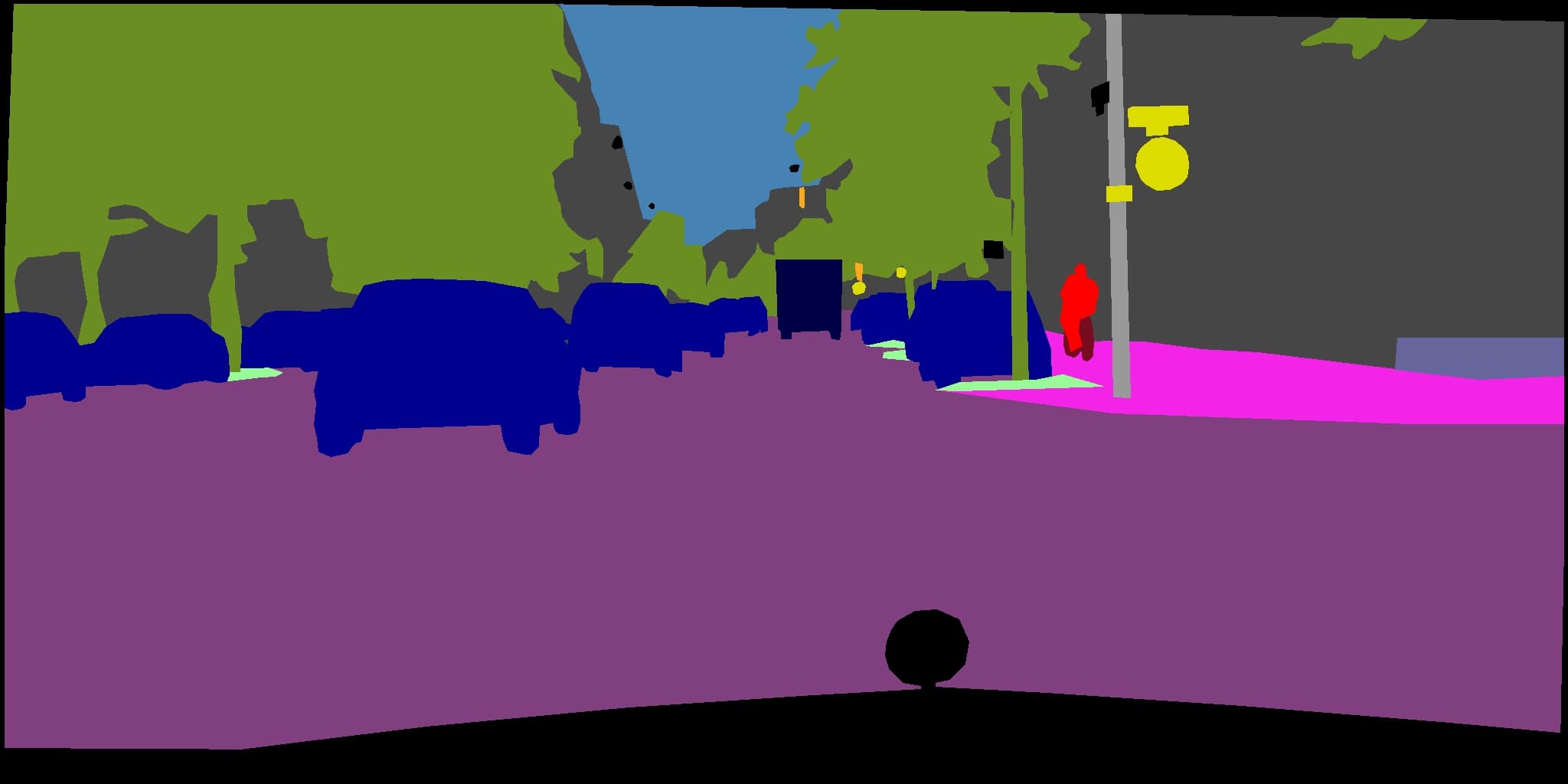}&
\includegraphics[width=.22\linewidth]{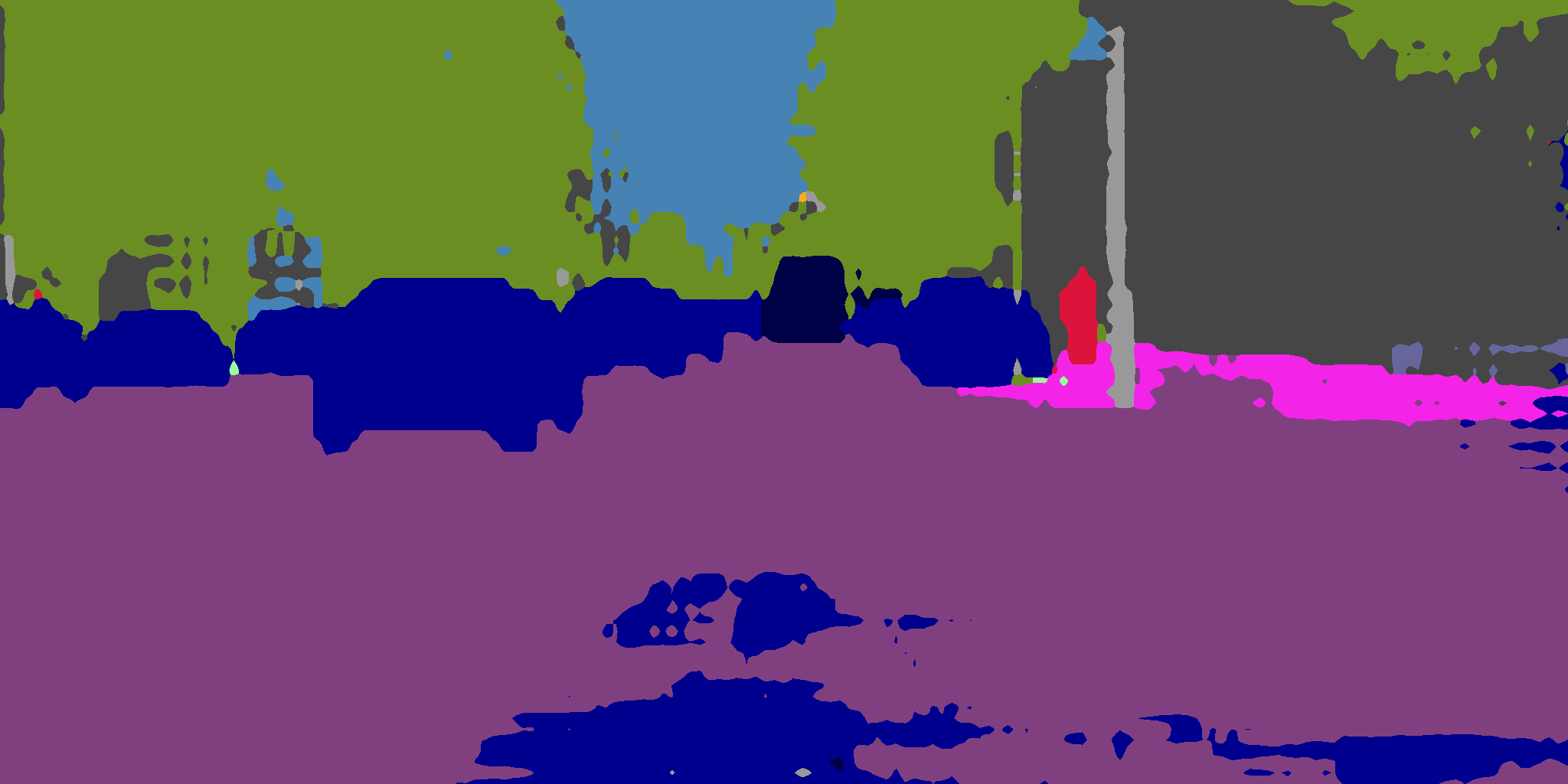}&
\includegraphics[width=.22\linewidth]{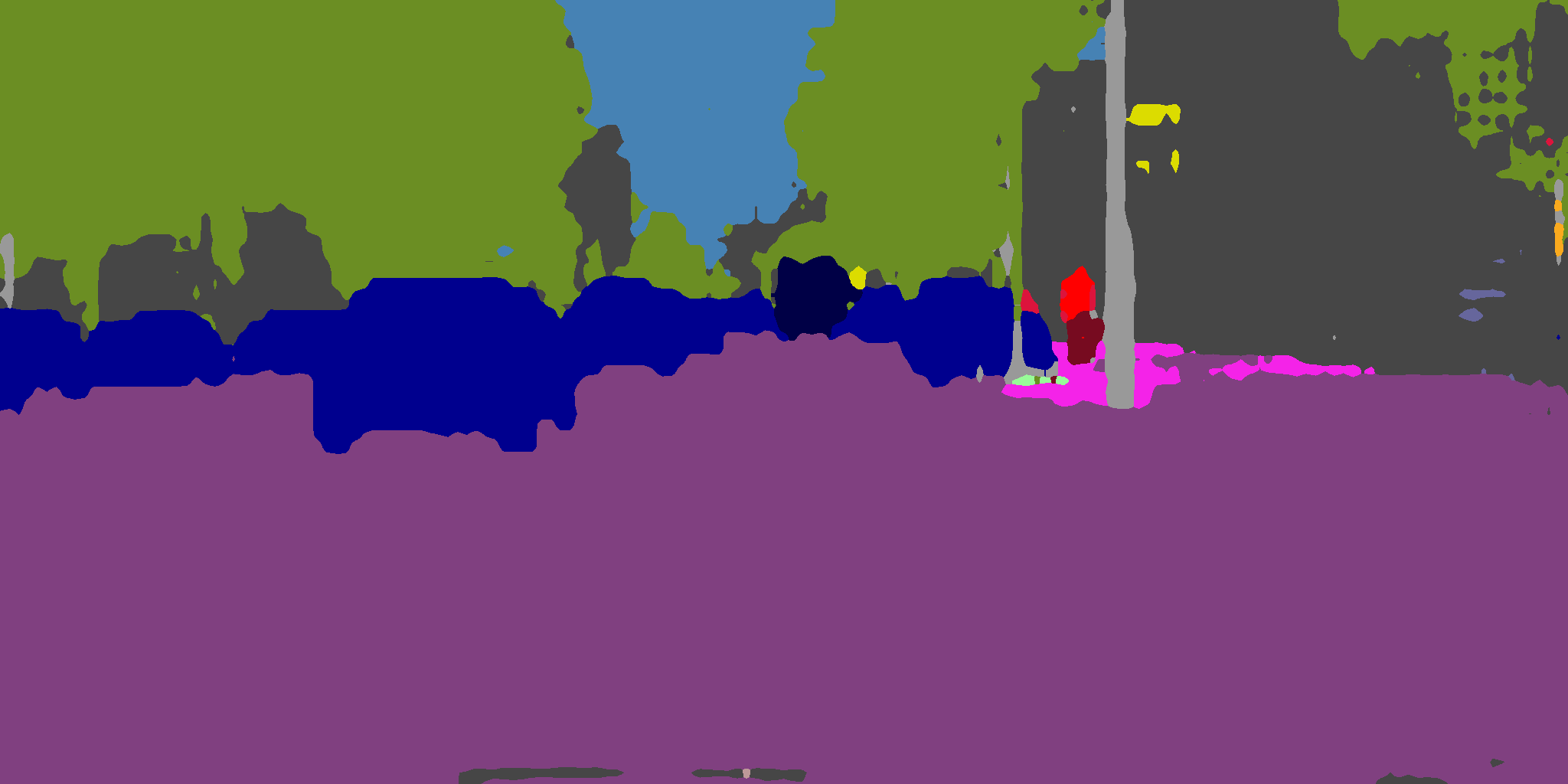}\\
\includegraphics[width=.22\linewidth]{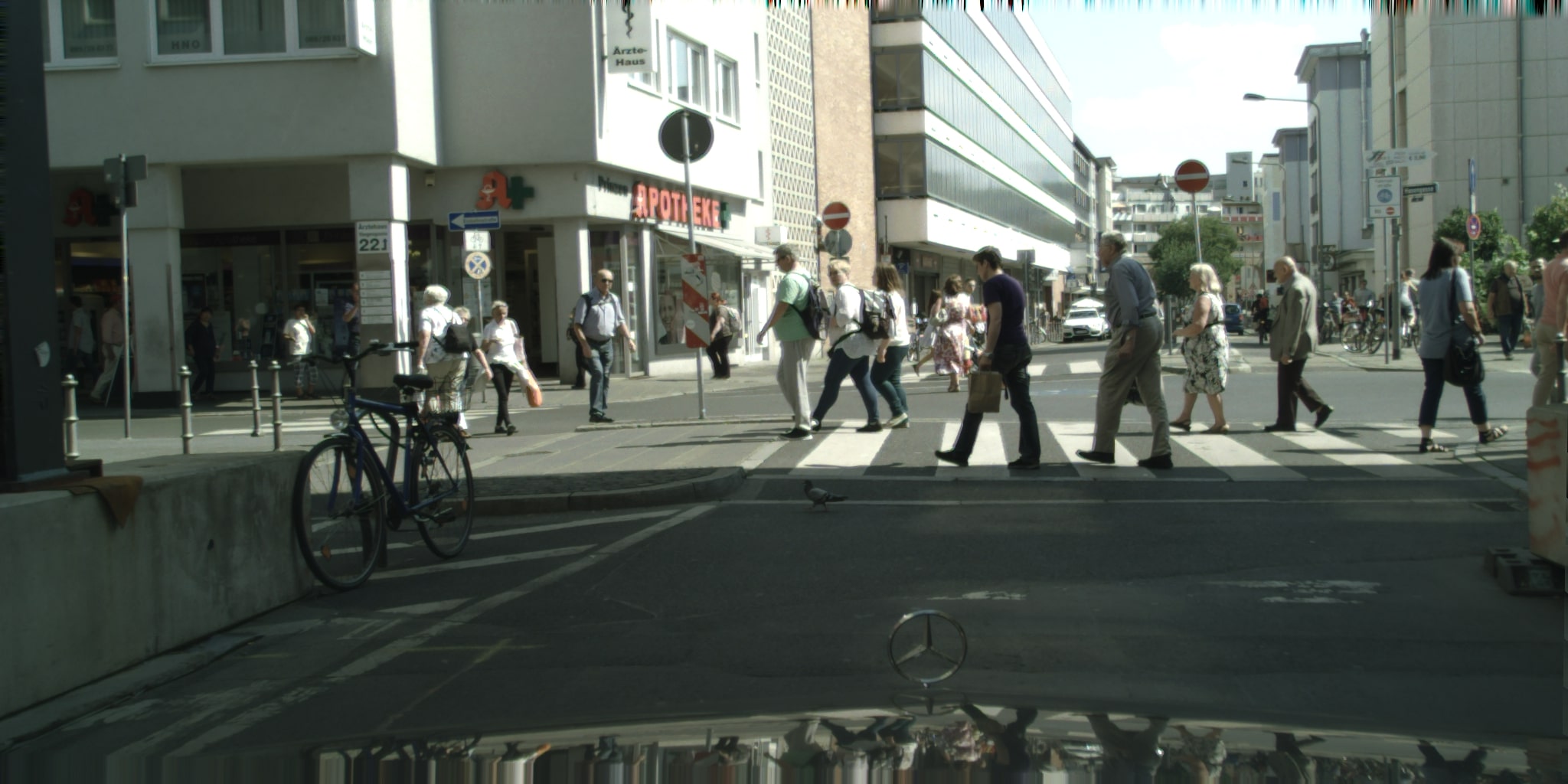}&
\includegraphics[width=.22\linewidth]{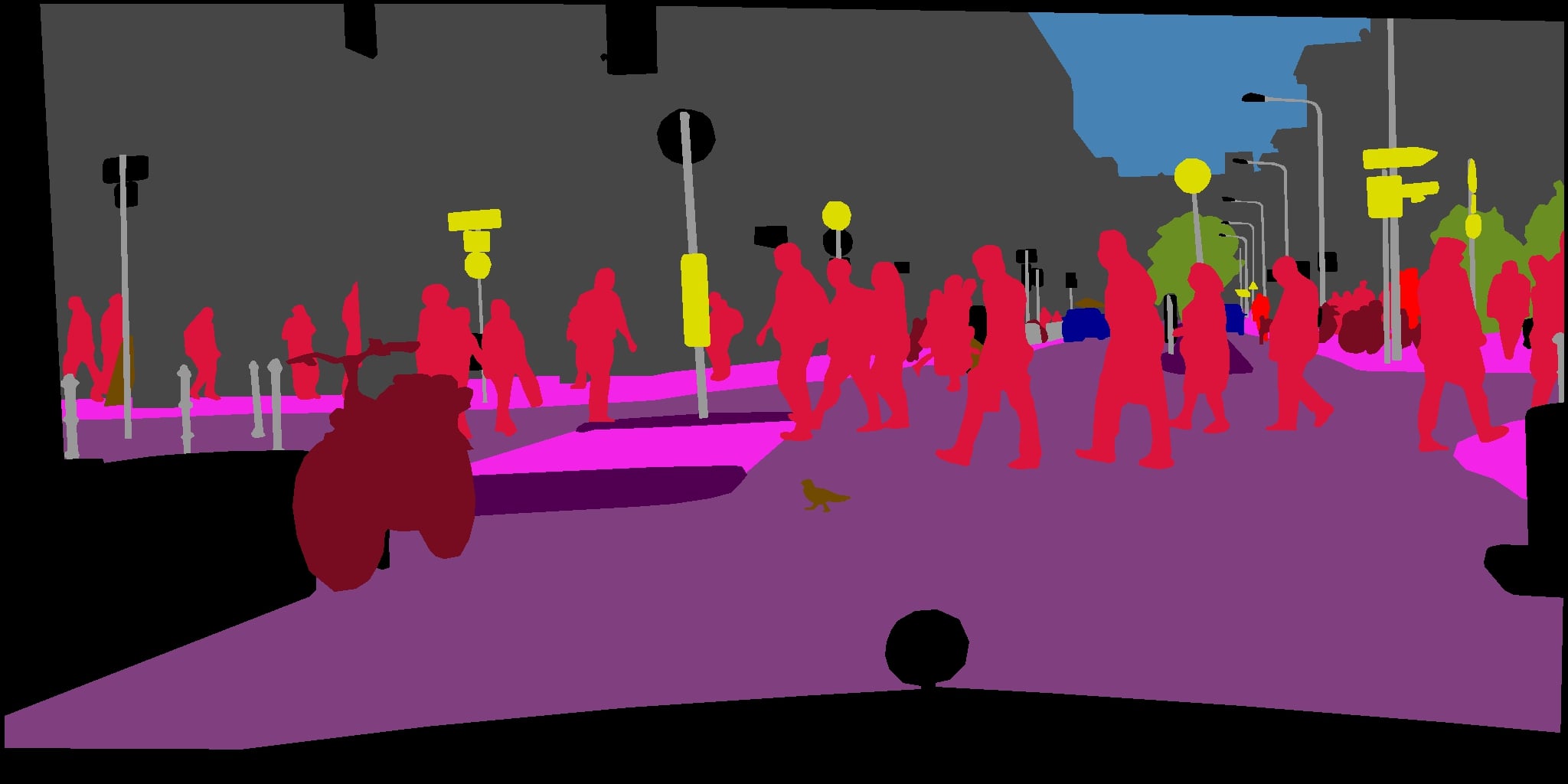}&
\includegraphics[width=.22\linewidth]{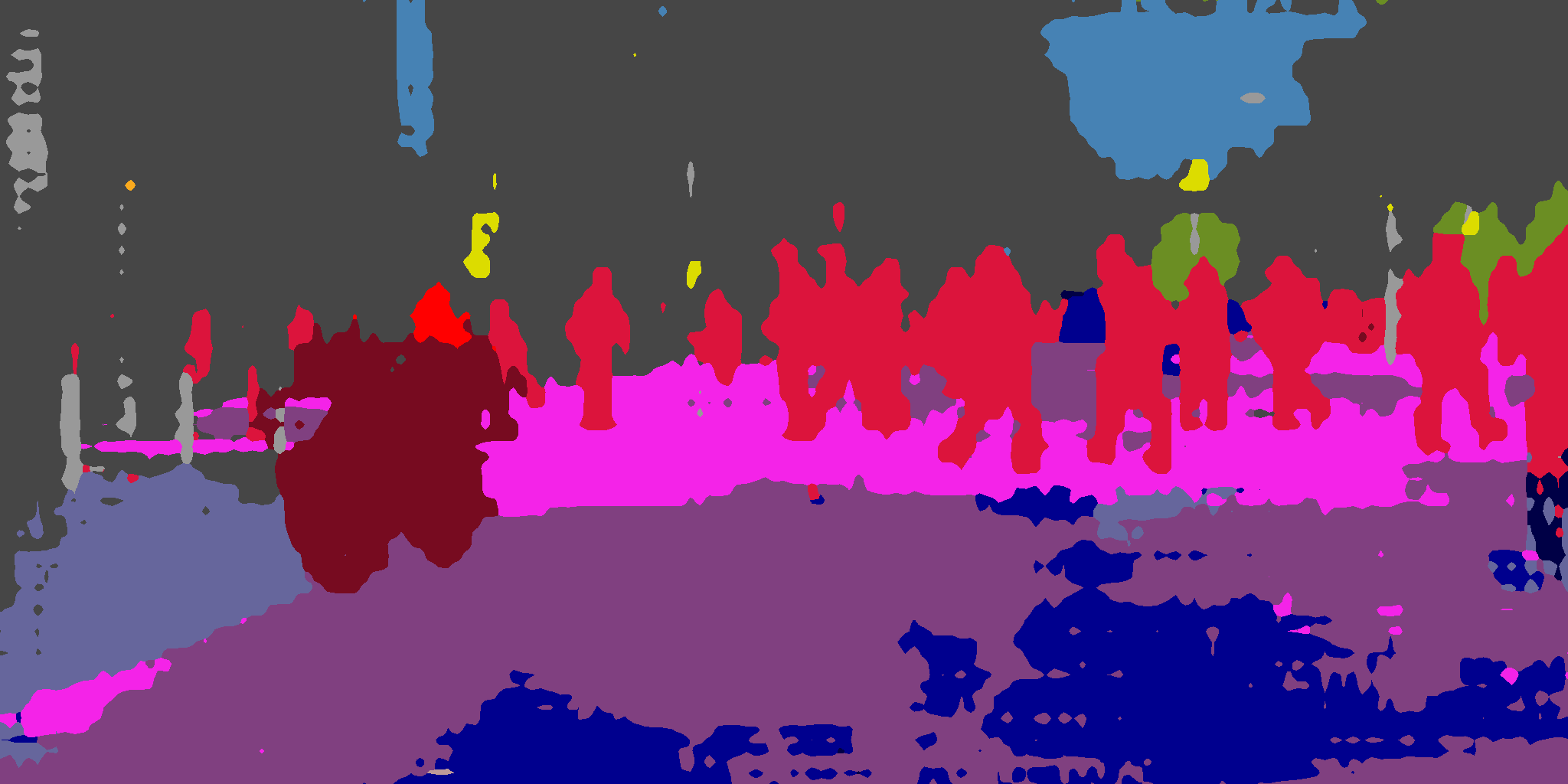}&
\includegraphics[width=.22\linewidth]{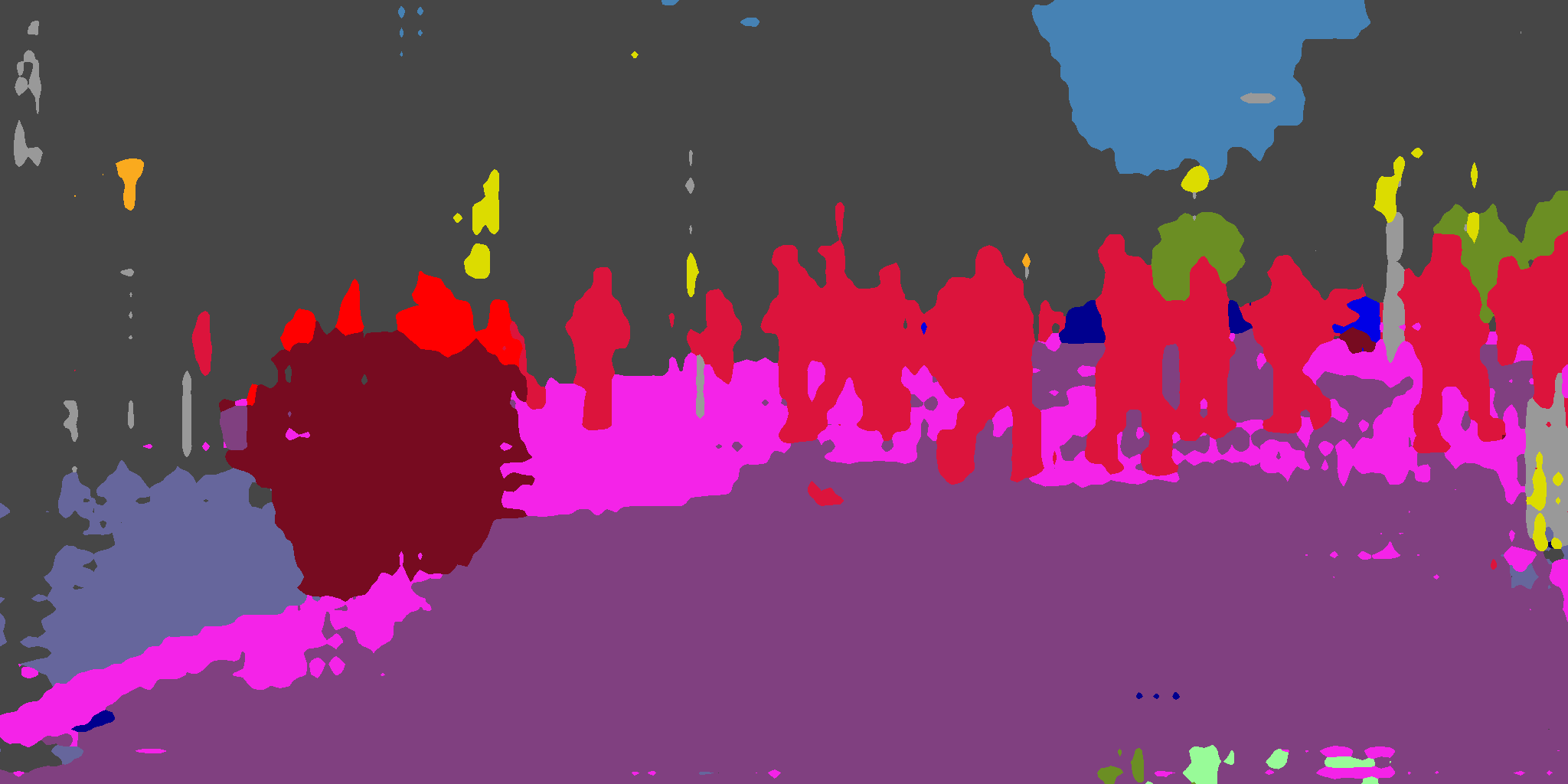}\\
\includegraphics[width=.22\linewidth]{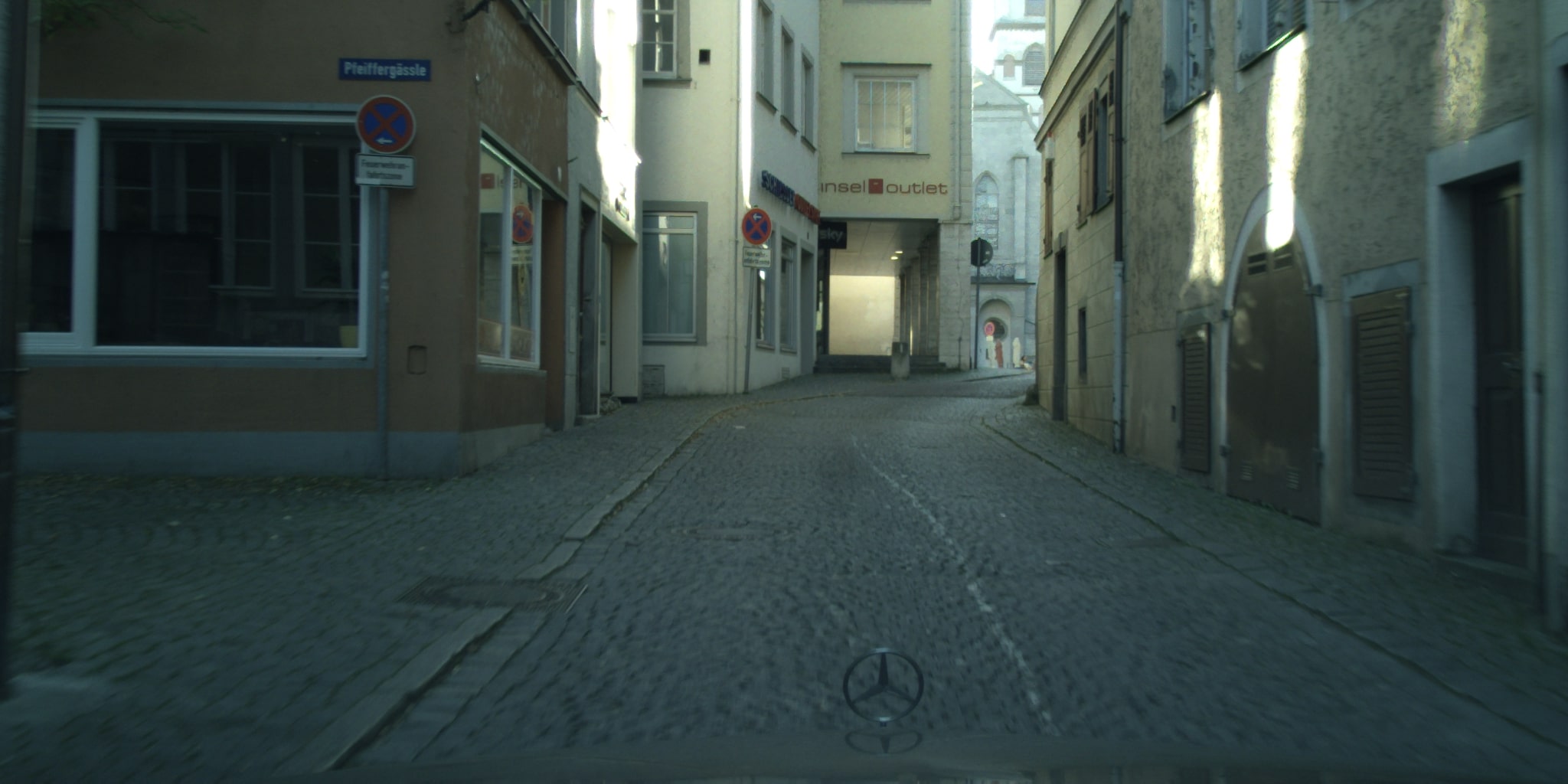}&
\includegraphics[width=.22\linewidth]{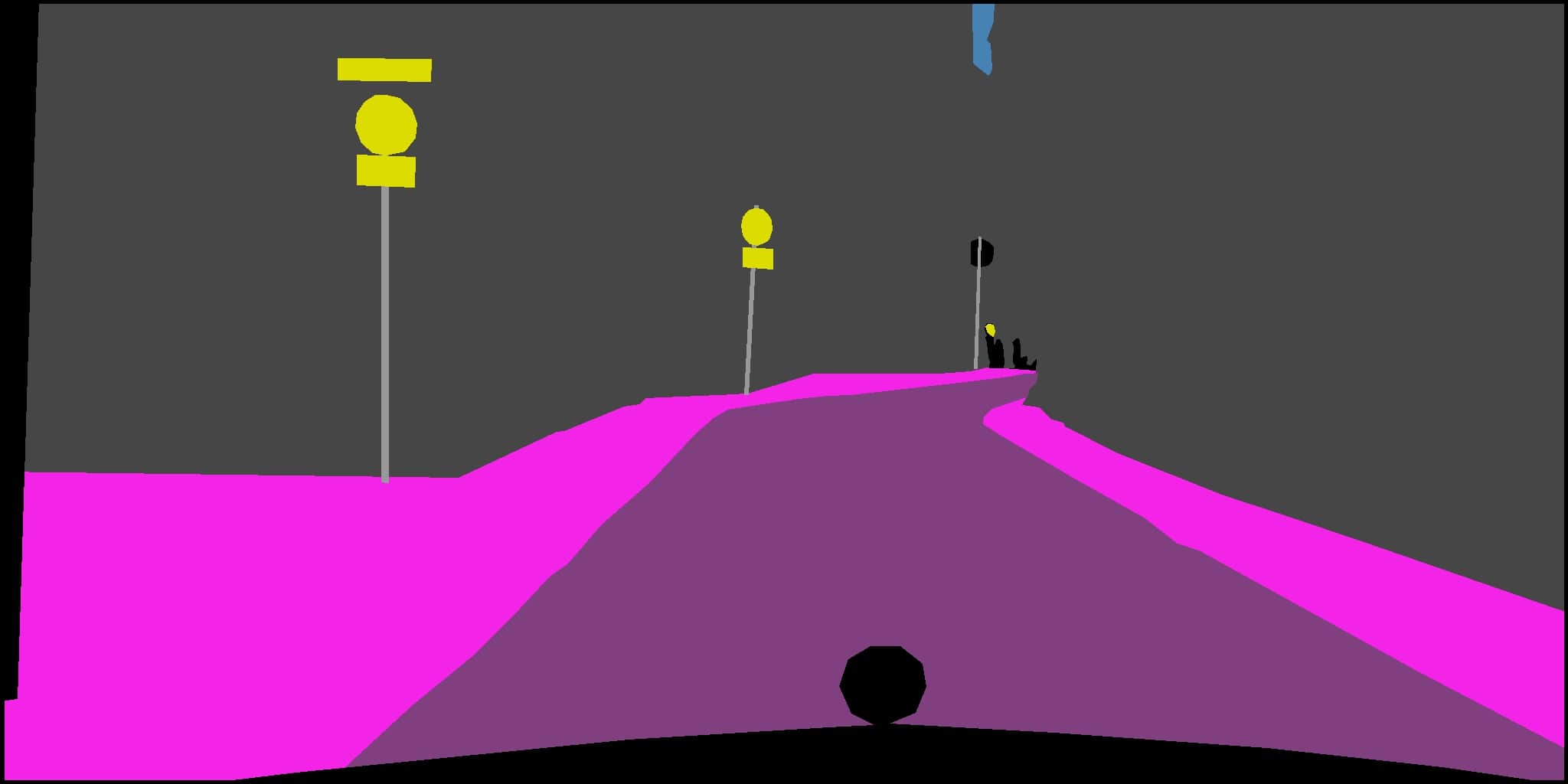}&
\includegraphics[width=.22\linewidth]{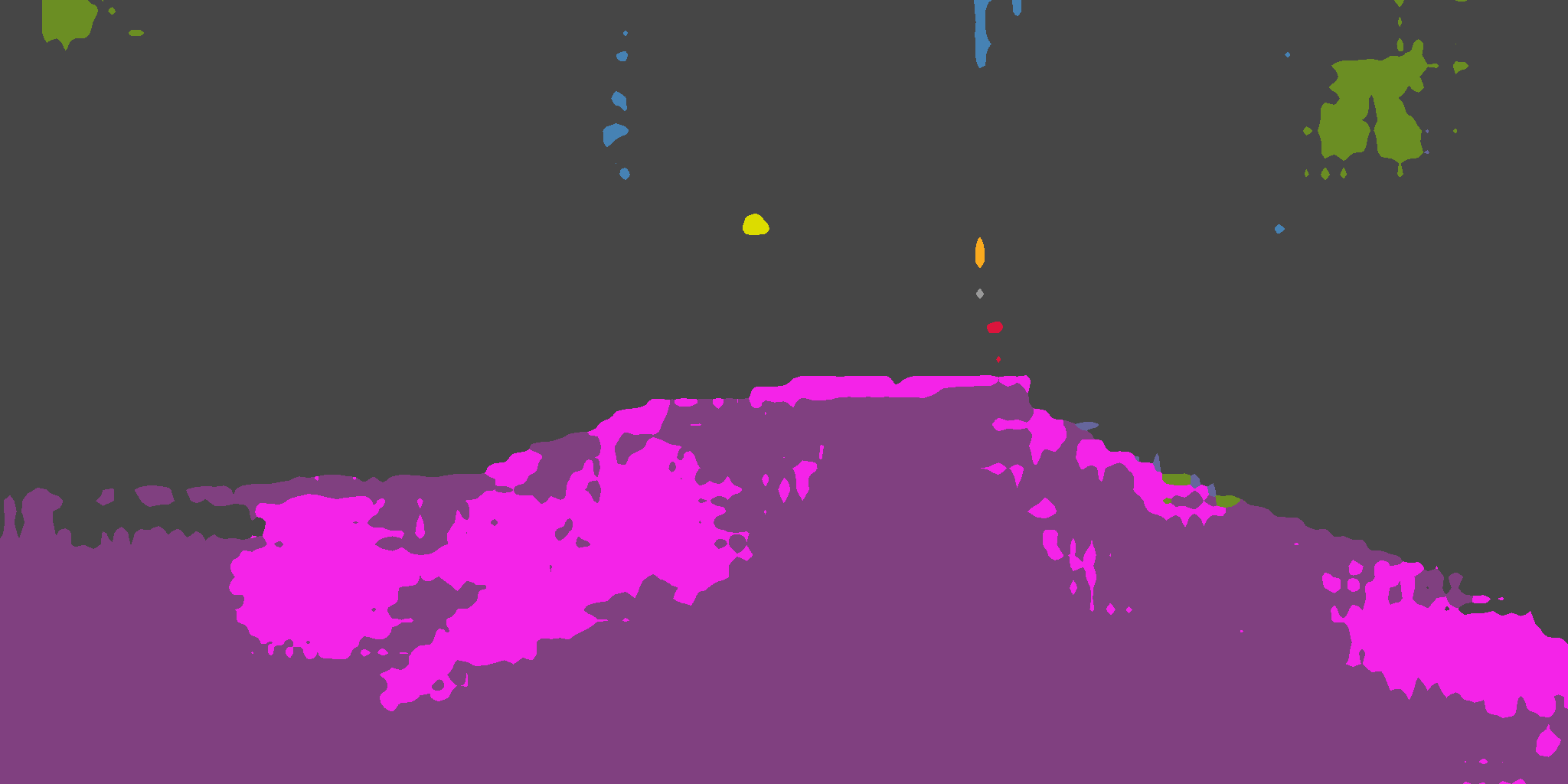}&
\includegraphics[width=.22\linewidth]{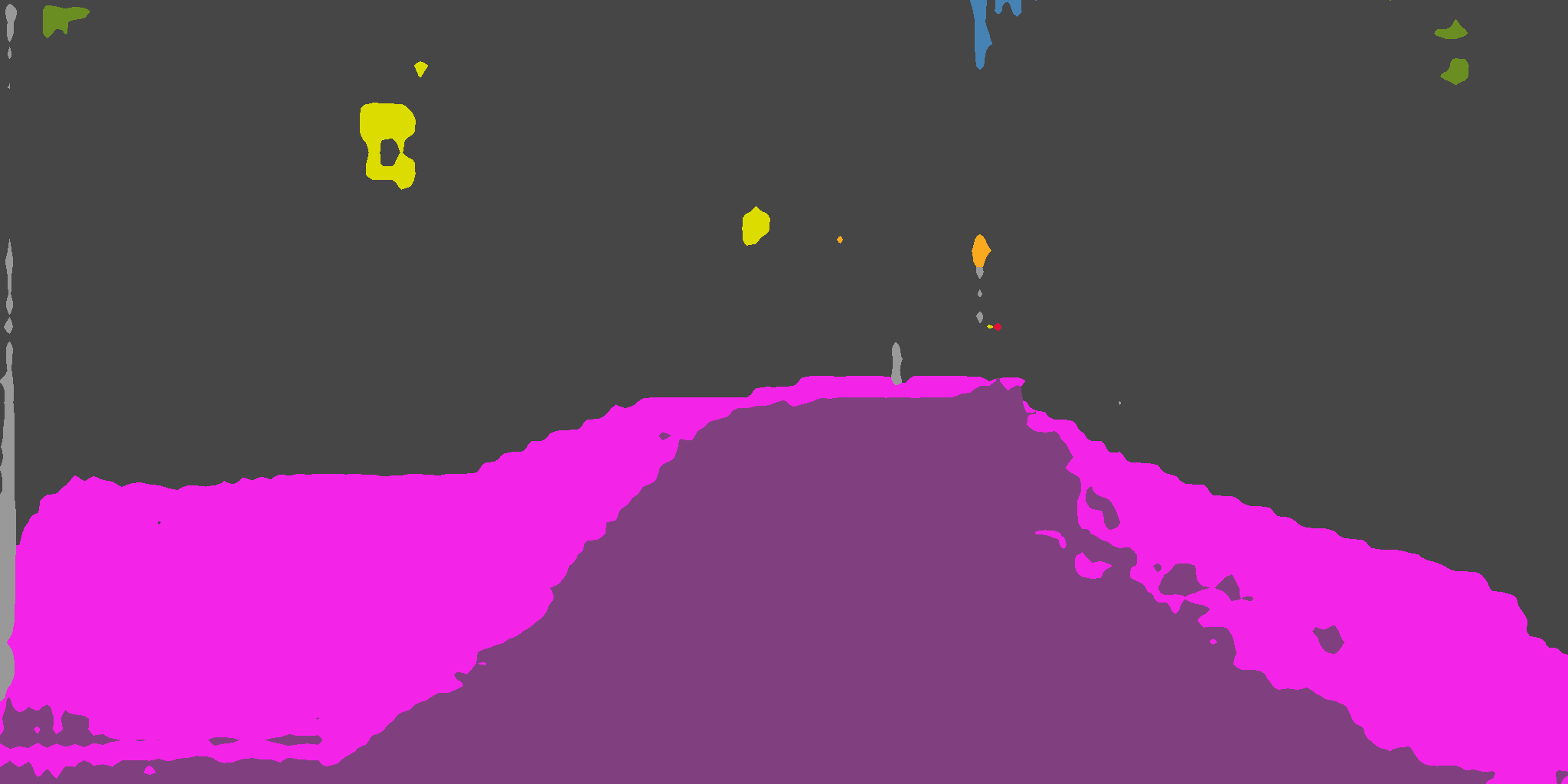}\\
\footnotesize{(a) Image}&
\footnotesize{(b) GT}&
\footnotesize{(c) FDA}&
\footnotesize{(d) \ours{}}
\end{tabular}
  \vspace{-3mm}
  \caption{\small {\bf Qualitative results.} (a) CityScapes images. (b) Ground-truth semantic segmentations. (c) FDA~\cite{Yang20a} results given 50 labeled target domain images. (d) Our results given the same 50 labeled target domain images. Our semantics maps tend to be smoother and to preserve the scene structure better. Note, for example, the building in the first and fourth row; the road in the second, third, and fourth row; and the sidewalk in the fifth row.}
  \label{fig:qualitative}
  \end{figure*}

\subsection{Datasets}

To compare these baselines against our approach, we use one real-world dataset and two synthetic ones and follow the same protocols as in these earlier methods for a fair comparison. They are:

\parag{CityScapes~\cite{Cordts16}} is a real-world benchmark dataset for semantic segmentation featuring driving scenes and dense manual annotations.  We use the $2,975$ images from the training set as the target domain for training.  We test on the $500$ validation images and resize the images to $1024 \times 512$ without any random cropping, as in~\cite{Yang20b,Yang20a}.

\parag{GTA5~\cite{Richter16}} contains $24,966$ synthetic images captured from a video game.  As in~\cite{Yang20b,Yang20a}, we resize the images to $1280 \times 720$ and randomly crop them to $1024 \times 512$ during training.  The original dataset features $33$ different categories of pixel-wise semantic labels but we only use the $19$ classes that are shared with CityScapes as on our baselines~\cite{Vu19,Wang20,Yang20a}.

\parag{SYNTHIA~\cite{Ros16}} is also a synthetic dataset.  As others~\cite{Vu19,Wang20,Yang20a}, we use the SYNTHIA-RAND-CITYSCAPES subset which comprises $9,400$ annotated images. During training, we also randomly crop the image to $1024 \times 512$ and evaluate with $13$  classes to follow the standard protocol.

\parag{} We use CityScapes as the target domain and either GTA5 or SYNTHIA as the source domain. We refer to the two resulting tasks as GTA5$\rightarrow$CityScapes and  SYNTHIA$\rightarrow$CityScapes. We use all the labels from the source domain and either none for the target domain or only a handful, which we define as the unsupervised and semi-supervised cases, respectively. 



\begin{table*}[t]
  \centering
    \scalebox{0.73}{
    \begin{tabular}{ccccccccccccccccccccc|c}
  \toprule
 \rotatebox{90}{\# labeled} & \rotatebox{90}{Method} & \rotatebox{90}{road} & \rotatebox{90}{sidewalk}& \rotatebox{90}{building} & \rotatebox{90}{wall}& \rotatebox{90}{fence} & \rotatebox{90}{pole}& \rotatebox{90}{light} & \rotatebox{90}{sign}& \rotatebox{90}{vegetation} & \rotatebox{90}{terrain}& \rotatebox{90}{sky} & \rotatebox{90}{person}& \rotatebox{90}{rider} & \rotatebox{90}{car}& \rotatebox{90}{truck} & \rotatebox{90}{bus}& \rotatebox{90}{train} & \rotatebox{90}{motocycle}& \rotatebox{90}{bicycle} & mIoU \\
  \midrule
    \multirow{3}{*}{\rotatebox{90}{0}} 
&MinEnt~\cite{Vu19} & 84.4& 18.7& 80.6& 23.8& 23.2& 28.4& 36.9& 23.4& 83.2& 25.2& {\bf 79.4}& 59.0& 29.9& 78.5& 33.7& 29.6& 1.7& {\bf 29.9}& 33.6& 42.30 \\
&AdvEnt~\cite{Vu19}  &  89.9 & 36.5 & 81.6 & {\bf 29.2} & 25.2 & 28.5 & 32.3 & 22.4 & {\bf 83.9} & 34.0 & 77.1 & 57.4 & 27.9 & 83.7 & 29.4 & 39.1 & 1.5 & 28.4 & 23.3 & 43.80   \\
&FDA~\cite{Yang20a} &  92.5 &  53.3 &  82.4& 26.5&  27.6& 36.4& 40.6& 38.9& 82.3& {\bf 39.8}& 78.0& {\bf 62.6}& 34.4& {\bf 84.9}&  34.1 & 53.1 & 16.9 & 27.7& 46.4 & 50.45 \\
&\ours{} & {\bf 93.3} & {\bf 54.2} &  {\bf 83.0} & 25.9& {\bf 28.1} &  {\bf 37.2}& {\bf 41.1} & {\bf 39.3} & 83.1&  38.9 & 78.2&  61.3 & {\bf 36.2} &  84.2 &  {\bf 35.8} &  {\bf 54.0} & {\bf 18.1} &  26.7 & {\bf 47.5} & {\bf 50.85}    \\
  \midrule
  \multirow{5}{*}{\rotatebox{90}{50}} 
&MinEnt~\cite{Vu19} & 92.6 & 56.8 &82.5 & 27.5 & 27.0 & 36.6 & 29.7 & 36.8 & 84.6 & 36.7 &78.2 & 60.5 &19.4 & 81.1 & 36.4 & 40.0 & 4.6 &19.4 & 52.3 & 47.51 \\
&AdvEnt~\cite{Vu19}  &  90.1 &  43.6 &  81.0 &  23.2 & 27.4 & 34.9 &27.5 &  28.0 & 84.6 & 26.6 & 75.7 & 60.9 &24.1 &  76.8 &39.1 &42.7 &  8.3 & 17.6 & 41.9 & 44.94   \\
&FDA~\cite{Yang20a} & 93.7& 59.6& 84.3&  {\bf 31.7}&  31.0&   37.0& 36.5&  42.1& 86.5&  {\bf 44.8}& 80.9&   {\bf 63.7}& {\bf 31.9}&  86.7& 47.2& 53.2 & {\bf 13.2} &33.8 & 50.4 & 53.10  \\
&ASS~\cite{Wang20} &  94.3 & 63.0 & 84.5 & 26.8 & 28.0 & 38.4 & 35.5 & 48.7 & {\bf 87.1} & 39.2& {\bf 88.8}& 62.2& 16.3& 87.6& 23.2& 39.2& 7.2& 24.4& {\bf 58.1}& 50.10    \\
&Universal~\cite{Kalluri19} &  91.2 &  44.1 &  82.4 &  25.0 & 29.2 & 36.1 &29.0 &  32.5 & 81.9 & 29.4 & 77.1 & 64.0 &30.2 &  73.5 &45.0 &45.9 &  11.2 & 18.2 & 44.6 & 46.87 \\
&\ours{} & {\bf 94.4}&  {\bf 63.6} & {\bf 85.3} & 26.4 &{\bf 31.3} &{\bf 40.3} & {\bf 41.5} & {\bf 53.0} &87.1 & 43.4 &85.4 &62.9 & 30.1 & {\bf 88.0} &{\bf 48.3} & {\bf 55.2} & 8.3 & {\bf 34.0} &51.0 & {\bf 54.17}   \\
  \midrule
  \multirow{5}{*}{\rotatebox{90}{100}} 
&MinEnt~\cite{Vu19} & 93.0& 56.3& 83.2& 27.5& 26.0& 37.1 & 31.9 & 40.4 & 85.2 & 42.5 & 82.3&  60.6&  27.3&  83.7& 41.8& 45.3& 1.6& 20.9& 45.1& 49.02 \\
&AdvEnt~\cite{Vu19}  &  91.3&  51.0&   82.2&  23.2&  26.1& 37.3& 32.3&  33.0& 84.3&  34.3&   77.0&  61.7& 30.0&  84.0&  39.3& 42.3&  0.3& 16.2& 45.1& 46.89 \\
&FDA~\cite{Yang20a} &  94.4 &62.2 &85.2 &{\bf 32.7} &32.6 &38.6 &39.5 &47.6 &86.8 & 48.9 & 85.1 & 64.5 & 36.8 &  87.5 & 46.1 & {\bf 53.3} & 0.7 &33.0 & 52.4 & 54.10  \\
&ASS~\cite{Wang20} & {\bf 96.0}& {\bf 71.7}& 85.9& 27.9& 27.6& {\bf 42.8}& {\bf 44.7}& {\bf 55.9}& {\bf 87.7}& 46.9& {\bf 89.0}& {\bf 66.0}& 36.4& 88.4& 28.9& 21.4& {\bf 11.4}& 38.0& {\bf 63.2}& 54.20   \\
&Universal~\cite{Kalluri19} & 92.2& 57.4& 84.1& 28.7& 22.8& 41.2 & 33.0 & 41.7 & 83.9 & 45.8 & 80.7& 61.1&  29.4 &  80.8& 41.5 & 42.9& 6.2& 11.5& 49.1 & 49.16 \\
&\ours{} & 94.6 &  66.1 &  {\bf 86.3} &  26.2 & {\bf 34.9} &  38.9 & 40.5 &  52.9 & 87.3 & {\bf 49.2} &  87.9 &64.7 &  {\bf 36.8} & {\bf 88.6} & {\bf 48.4} &  45.0 & 2.8 & {\bf 38.8} &  58.5 & {\bf 55.17}   \\
  \midrule
  \multirow{5}{*}{\rotatebox{90}{200}} 
&MinEnt~\cite{Vu19} & 93.9& 59.0&  84.7& 22.4& 30.2& 38.4& 36.2& 47.3& 86.0& 42.4& 82.4& 62.1& 33.0& 87.1& 44.2&  39.0& {\bf 28.5}& 20.3& 52.3& 52.07 \\
&AdvEnt~\cite{Vu19}  & 92.7&   58.4&  84.0&  22.4& 30.1 &  36.3 & 37.3&  43.4 & 85.6 & 35.7 &  83.4 & 61.3 &  28.0 &  80.9 & 39.5 &41.6 & 22.9 & 19.0 &  51.7 & 50.23   \\
&FDA~\cite{Yang20a} &   94.7 &62.8 & 85.9 &  {\bf 32.9} &  31.9 &  38.5 &  41.9 &  51.7 & 86.9 & 47.5 & 85.2 & 64.8 &  36.7 & 87.9 &  48.6 & {\bf 58.2} & 19.2 & 37.0 & 54.9 & 56.16   \\
&ASS~\cite{Wang20} & {\bf 96.1}& {\bf 71.9}& 85.8& 28.4& 29.8& {\bf 42.5}& 45.0& {\bf 56.2}& {\bf 87.4}& 45.0& {\bf 88.7}& 65.8& 38.2& {\bf 89.6}& 42.2& 35.9& 17.1& 35.8& {\bf 61.6}& 56.00    \\
&Universal~\cite{Kalluri19} & 93.1& 64.8&  82.1& 21.5& 25.8& 39.0& 39.8& 43.4& 82.7& 47.9& 83.8& 63.6& 35.8 & 83.2& {\bf 49.0} &  38.1&  19.8& 24.9& 50.4 & 52.04 \\
&\ours{} &  94.2 & 62.2& {\bf 85.9} & 29.0& {\bf 35.0}&  41.7&{\bf 45.7}& 55.4&87.4& {\bf 49.1}& 86.1 & {\bf 65.8} & {\bf 40.2} &  88.8& 47.6& 58.1&12.7& {\bf 40.2}& 57.2 & {\bf 56.96}  \\
  \midrule
  \multirow{5}{*}{\rotatebox{90}{500}} 
&MinEnt~\cite{Vu19} & 94.8 & 65.6 & 86.0 & 32.1 & 37.3 & 40.7 & 40.5 & 53.7 &  86.7 &  44.8 &  86.3 & 63.2 & 31.0 &  86.3 & 49.8 & 53.6 &  13.1 &  26.7 &  58.1 &  55.26 \\
&AdvEnt~\cite{Vu19}  & 94.9 &  67.5 &  85.7 & 28.4 & 36.3 &41.3 & 40.8 &  53.1 & 87.3 & 45.6 & 84.4 & 64.2 & 36.4 & 87.5 & 50.7 & 46.9 & 13.4 & 29.2 & 59.4 & 55.42   \\
&FDA~\cite{Yang20a} &  94.6 &  64.9 &  86.6 & {\bf 36.3} &   38.9 &  42.6 &  46.2 &  56.4 & 87.8 &  {\bf 52.4} &   85.9 & 65.9 &    39.4 &  89.3 &  56.4 &    62.0 &  {\bf 23.6} &  39.3 &  56.1 & 59.19   \\
&ASS~\cite{Wang20} & {\bf 96.2}& 72.7 & {\bf 87.6}& 35.1& 31.7& {\bf 46.6}& 46.9& {\bf 62.7}& {\bf 88.7}& 49.6& {\bf 90.5}& {\bf 69.2}& 42.7& {\bf 91.1}& 52.6& 60.9& 9.6& {\bf 43.1}& {\bf 65.6}& 60.20   \\
&Universal~\cite{Kalluri19} & 94.2 &  {\bf 72.8} &  84.9 & 29.9 & 38.5 &43.5 & 42.7 &  55.4 & 85.8 & 48.0 & 83.5 & 66.1 & 39.2 & 86.4 & 51.8 & 48.5 & 16.9 & 31.5 & 61.1 & 56.88 \\
&\ours{} & 94.7 & 66.7 & 87.4 & 31.5 &{\bf 42.5} &42.2 & {\bf 47.0} & 58.5 &87.7 & 49.7 & 88.5 &68.1 &{\bf 44.5} &  89.5 & {\bf 62.3} &  {\bf 62.1} & 19.7 &  42.6 &   63.2 & {\bf 60.43}   \\
  \midrule
  \multirow{5}{*}{\rotatebox{90}{1000}} 
&MinEnt~\cite{Vu19} &  95.5 &  70.4 &   86.7 & 33.3 & 35.7 &  41.6 &  44.4 &  55.9 &  87.3 &  46.1 &  87.5 & 64.8 & 38.3 & 88.2 & 45.8 & 59.4 & 34.0 & 34.0 & 60.6 & 58.40 \\
&AdvEnt~\cite{Vu19}  &  95.4 & 70.1& 86.3& 35.9& 38.7& 40.5& 43.7&  55.4& 87.7& 50.5& 87.1& 65.6&  40.6&  87.9& 56.9& 51.0&  10.3&  36.7&  61.1&  57.98   \\
&FDA~\cite{Yang20a} & 96.0 & 71.9 &  87.2 & 31.7 & 39.7 & 44.0 & 47.5 & 59.1 & 88.0 & 51.1 &     88.8 &  69.4 &  47.8 &  89.9 & 63.0 & 67.2 & 36.6 & 45.9 &  60.3 &62.37  \\
&ASS~\cite{Wang20} & {\bf 96.8}& {\bf 76.3}& {\bf 88.5}& 30.5& 41.7& {\bf 46.5}& 51.3& {\bf 64.3}& {\bf 89.1}& {\bf 54.2}& {\bf 91.0}& 70.7& 48.7& {\bf 91.6}& 59.9& 68.0& {\bf 40.8}& {\bf 48.0}&  67.0 & 64.50   \\
&Universal~\cite{Kalluri19} &  95.2 &  74.3 &   87.6 & 35.9 & 37.2 &  43.8 &  45.2 &  53.1 &  85.4 &  49.8 &  84.3 & 68.3 & 41.0 & 86.5 & 48.3 & 62.2 & 37.5 & 38.6 & {\bf 67.5} & 60.09 \\
&\ours{} &  96.1 & 72.5 &  88.5 & {\bf 38.9} &  {\bf 47.7} &  45.8 &  {\bf 51.6} &   61.7 & 88.9 &   50.9 &   89.1 &  {\bf 71.4} &  {\bf 51.0} &   91.3 &  {\bf 68.1} &  {\bf 69.4} &  32.5 &  46.2 & 66.4 & {\bf 64.62}    \\
  \bottomrule
  \end{tabular}
  }
  \vspace{-2mm}
  \caption{\small {\bf Results on GTA5$\rightarrow$CityScapes:} We evaluate on the $19$ common classes shared between these two dataset. Our method is consistently better compared to baselines, particularly for few number of labeled target domain images.}
  \label{tab:eva_gta5}
\end{table*}

\begin{table}[t]\centering
  \scalebox{0.83}{
  \begin{tabular}{l cccccc}
  \toprule
                         & \multicolumn{6}{c}{mIoU on \# Labeled}                                      \\
  Method                 & 0          & 50         & 100        & 200        & 500        & 1000       \\
  \midrule
  MinEnt~\cite{Vu19}     & 44.20      & 52.86      & 56.41      & 57.93      & 62.54      & 66.04      \\
  AdvEnt~\cite{Vu19}     & 47.60      & 51.41      & 55.20      & 59.64      & 62.58      & 66.82      \\
  FDA~\cite{Yang20a} & 52.50 & 58.48      & 62.03      & 64.40      & 66.81      & 70.17      \\
  ASS~\cite{Wang20}      & -          & 60.70      & 62.10      & 64.80      & 69.80      & 73.00      \\
    Universal~\cite{Kalluri19}  & - & 53.63      & 57.05      & 60.26  & 63.41  &  68.47\\
  \ours{}                & {\bf 53.26}          & \bf{61.18} & \bf{63.39} & \bf{65.23} & \bf{70.26} & \bf{73.09} \\
  \bottomrule
  \end{tabular}
  }
  \vspace{-2mm}
  \caption{\small {\bf Results on SYNTHIA$\rightarrow$CityScapes:} We report mIoU over $13$ classes, following the standard protocol. Note that our method consistently outperforms others. Per-class results can be found in supplemental material.}
  \label{tab:eva_synthia}
\end{table}

\subsection{Comparative Results}
\label{sec:comparative}

In this section, we compare our results against the baselines and  report the results in Tables~\ref{tab:eva_gta5} and~\ref{tab:eva_synthia} as a function of the number of annotated images in the target domain. $0$ corresponds to the UDA setting while $50$, $100$, $200$, $500$ and $1000$ denote SSDA as in~\cite{Wang20}.
We provide qualitative results in Fig.~\ref{fig:qualitative}. 

Our approach outperforms others in overall mIoU and in most individual categories. ASS~\cite{Wang20} delivers the best performance in some categories but still does worse than our proposed method overall. Note that the performance gap is highest in the SSDA setting where we use only a small number of annotated target domain samples, such as 50. This is significant for many practical applications: Annotating 50 images is typically possible and therefore worth doing given the boost it provides. 

Moreover, the ASS approach is complementary to ours and both could be used jointly. The adversarial loss of ASS could be used as an extra loss term in our approach, which is something worth exploring in future work.

\subsection{Further Weakening the Annotations}

To further drive down the annotation cost and increase its practical appeal, we not only restrict the number of annotated images in the target domain but also perform only partial annotations within these images. 

To this end, we split the 200 labeled target domain images into $10 \times 10$ pixel blocks as in~\cite{Lin19} and, instead of annotating all of them, we randomly annotate only a subset and fill the others with pseudo labels.


\begin{table*}
  \begin{subtable}[b]{0.2\textwidth}\centering
    \scalebox{0.8}{
    \begin{tabular}{cc}
    \toprule
    Annotation & mIoU \\
    \midrule
    $100\%$ & {\bf 56.96}  \\
    \midrule
    $75\%$ & 56.68  \\
    $50\%$ & 56.21 \\
    $25\%$ &55.94 \\
    \bottomrule
    \end{tabular}
    }
    \caption{}
    \label{tab:abl_weak_anno}
  \end{subtable}
  \begin{subtable}[b]{0.23\textwidth}\centering
    \scalebox{0.8}{
    \begin{tabular}{cc}
    \toprule
    Pseudo Label Quality  & mIoU \\
    \midrule
    Low  & 56.48\\
    Median & 56.96 \\
    High & {\bf 58.05}\\
    \bottomrule
    \end{tabular}
    }
    \caption{}
    \label{tab:abl_sens}
  \end{subtable}
  \begin{subtable}[b]{0.18\textwidth}\centering
    \scalebox{0.8}{
    \begin{tabular}{cc}
    \toprule
    $\tau$ & mIoU \\
    \midrule
    $0.05$ & 53.43\\
    $0.07$ & {\bf 54.17}\\
    $0.1$ & 53.95\\
    \bottomrule
    \end{tabular}
    }
    \caption{}
    \label{tab:abl_tau}
  \end{subtable}
  \begin{subtable}[b]{0.18\textwidth}\centering
    \scalebox{0.8}{
    \begin{tabular}{ccc}
    \toprule
    $\alpha$ & $\beta$  & mIoU \\
    \midrule
    $1$ & $80$     & 53.74  \\
    $3$ & $70$     &  {\bf 54.17} \\
    $10$ & $40$     &  53.85 \\
    \bottomrule
    \end{tabular}
    }
    \caption{}
    \label{tab:abl_alpha_beta}
  \end{subtable}
  \begin{subtable}[b]{0.18\textwidth}\centering
    \scalebox{0.8}{
    \begin{tabular}{cc}
    \toprule
    Patch Size &  mIoU \\
    \midrule
    32 $\times$ 16  &  53.32\\
    64 $\times$ 32 & {\bf 54.17}  \\
    256 $\times$ 128 & 53.39  \\
    \bottomrule
    \end{tabular}
    }
    \caption{}
    \label{tab:abl_patch_size}
  \end{subtable}
  \\
  \begin{subtable}[b]{0.15\textwidth}\centering
    \scalebox{0.8}{
    \begin{tabular}{cc}
    \toprule
    Method &  mIoU \\
    \midrule
    Exact  &  53.25\\
    Pyramid& {\bf 54.17}  \\
    \bottomrule
    \end{tabular}
    }
    \caption{}
    \label{tab:abl_matching}
  \end{subtable}  
  \begin{subtable}[b]{0.23\textwidth}\centering
    \scalebox{0.8}{
    \begin{tabular}{ccc}
    \toprule
    $\lambda_{cont}^{GT}$ & $\lambda_{cont}^{PSE}$ & mIoU \\
    \midrule
    $0$       & $0$         &  53.06 \\
    $1e^{-3}$ & $0$         &  53.57 \\
    $1e^{-3}$ & $1e^{-3}$   &  53.15 \\
    $1e^{-3}$ & $1e^{-4}$   &  {\bf 54.17} \\
    $1e^{-4}$ & $1e^{-4}$   &  53.62 \\
    \bottomrule
    \end{tabular}
    }
    \caption{}
    \label{tab:abl_cont_loss_wgts}
  \end{subtable}
  \begin{subtable}[b]{0.40\textwidth}\centering
    \scalebox{0.8}{
    \begin{tabular}{lcc}
    \toprule
    Loss &  \shortstack{mIoU \\ {\bf w/} FDA}  & \shortstack{mIoU \\ {\bf w/o} FDA}\\
    \midrule
    $L_{base}$ & 51.93 &50.97\\
    $L_{base}+L_{self}$ & 53.06& 52.86\\
    $L_{base}+L_{self}+L_{cont}$ & {\bf 54.17}& {\bf 54.08}\\
    \bottomrule
    \end{tabular}
    }
    \caption{}
    \label{tab:abl_losses_impact}
  \end{subtable}
  \begin{subtable}[b]{0.15\textwidth}\centering
    \scalebox{0.8}{
    \begin{tabular}{lc}
    \toprule
    Method & mIoU \\
    \midrule
    \oursP{}&  52.04 \\
    \ours{}& {\bf 54.17} \\
    \bottomrule
    \end{tabular}
    }
    \caption{}
    \label{tab:abl_pretrain}
  \end{subtable}
  \vspace{-2mm}
  \caption{\small {\bf Ablation Study on:} (a) Weak annotation; (b) Quality of pseudo label; (c) Temperature parameter; (d) Threshold values for patch matching; (e) Patch size; (f) Matching strategy; (g) Weights of contrastive loss; (h) Contribution of each loss term; (i) Contrastive learning.}
  \label{tab:ablation}
\end{table*}

We then use our approach as described above. We report the results in
Tab.~\ref{tab:abl_weak_anno} as a function of the annotated percentage of each one of the 200 target domain images we use for this purpose. As can be seen in the table, we  can further cut the annotation cost in the target domain by $75\%$ with only a slight performance drop. 
Interestingly, comparing Tab.~\ref{tab:abl_weak_anno} and Tab.~\ref{tab:eva_gta5} shows that annotating $25\%$ of the blocks in 200 images delivers much better performance than fully annotating 50 images, while representing about the same annotation effort. 

\subsection{Ablation Study}
\label{sec:ablation}

\parag{Quality of Pseudo Labels.} 

To analyze how the quality of the pseudo labels affects our contrastive loss, we evaluate it using pseudo labels of different quality. Let the pseudo labels generated by the model trained using $100$, $200$, and $500$ annotated target domain images be the {\it low}, {\it median}, and {\it high} quality ones, respectively. We then use these labels to compute the contrastive loss but compute the other losses as if we had 200 annotated target domain images.  As shown in Tab.~\ref{tab:abl_sens}, the {\it high quality} labels give the best results, which confirms their importance.

\parag{Training Stability.} 

\begin{figure}[t]
\centering
\includegraphics[width=0.8\linewidth]{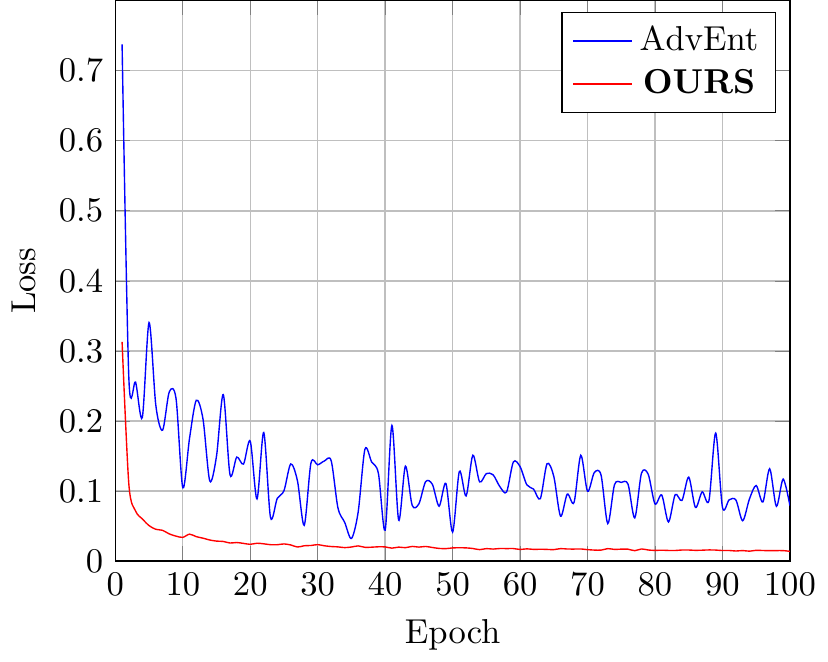}
\caption{\small {\bf Training curve.}  Our training curve is much smoother than the one of AdvEnt~\cite{Vu19} because our approach does not rely on adversarial learning. }
\label{fig:loss}
\end{figure}

To highlight that our approach is easier to train than methods based on adversarial learning, we compare the training loss between \ours{} and AdvEnt. As shown in Fig.~\ref{fig:loss}, our training curve is much better behaved.


\parag{Hyper-Parameters and Design Choices.} 

Finally, we demonstrate the impact of various hyper-parameter and design choices on our model with $50$ labeled real images.

Tables~\ref{tab:abl_tau}, \ref{tab:abl_alpha_beta} and \ref{tab:abl_patch_size} show the influence of hyper-parameters specific to the contrastive loss: The temperature parameter $\tau$, the thresholds $\alpha$ and $\beta$ that define positive and negatives pairs of patches, and the patch size itself.

In Tab.~\ref{tab:abl_matching}, we compare the pyramid matching algorithm of Section~\ref{sec:patch_similarity} against a simplified {\it Exact} matching scheme that computes the Hamming distance between the two flattened label patches. As expected, our more sophisticated scheme delivers better results. 

In Tab.~\ref{tab:abl_cont_loss_wgts}, we analyze the weighting of the contrastive loss terms applied on real and pseudo ground truth, respectively.  Even though both ground truth and pseudo-labels are useful, using higher weights $\lambda_{cont}^{GT}$ of Eq.~\ref{eq:loss_all} for the ground truth labels than $\lambda_{cont}^{PSE}$ for the pseudo labels is advisable.  

We analyze the impact of the individual loss terms of our model in Tab.~\ref{tab:abl_losses_impact} with and without using FDA-adapted input images.  All the proposed loss terms improve the performance in both cases.  Interestingly, the gain of using FDA becomes smaller when adding the contrastive loss, indicating its potential for domain adaptation.

Finally, we compare our training strategy and use of the contrastive loss against a pre-training strategy akin to that of~\cite{Chaitanya20}. In Tab.~\ref{tab:abl_pretrain}, \oursP{} is similar to \ours{} but uses the contrastive loss only for model pre-training as opposed to using it jointly with the semantic segmentation loss term during the training phrase as we normally do.  As the table shows, this is less effective.

\section{Conclusion}

We introduced a new domain adaptation algorithm for semantic segmentation. Our main contribution is a novel patch-wise contrastive loss that aligns image sub-regions across domains when they exhibit similar structures in label space. It enables our algorithm to outperform state-of-the-art methods both in unsupervised and semi-supervised scenarios. 

We have shown that our approach naturally extends to the  weakly-supervised case in which we annotate the images only partially. In future work, we will leverage this ability to implement an active learning scheme in which the image blocks to be annotated are chosen automatically. This should result in an even lower-cost and highly practical semi-supervised approach.

{\bf Acknowledgments}  This work was completed during an internship at Amazon Prime Air and supported in part by the Swiss National Science Foundation.

\newpage

{\small
\bibliographystyle{ieee_fullname}
\bibliography{short,string,egbib}
}

\newpage

\end{document}